\PassOptionsToPackage{table}{xcolor}

\documentclass[10pt,twocolumn,letterpaper,table]{article}
\usepackage{cvpr}              
\usepackage[linesnumbered,ruled]{algorithm2e}
\usepackage{todonotes}
\usepackage{nicematrix}
\usepackage{xcolor}

\definecolor{DarkMagenta}{rgb}{0.7, 0.0, 0.7}
\newcommand{\bo}[1]{#1}

\definecolor{DarkBlue}{rgb}{0.0, 0.0, 0.8}

\definecolor{DarkGreen}{rgb}{0.0, 0.5, 0.0}

\newcommand{\map}{f}
\newcommand{\layer}{\Phi}
\newcommand{\planlayer}{\Psi}
\newcommand{\liftlayer}{\tilde{\planlayer}}
\newcommand{\mesh}{\mathbf{M}}
\newcommand{\lap}{L}
\newcommand{\loss}{\mathcal{L}}
\newcommand{\lossarap}{\loss_{\text{elastic}}}
\newcommand{\losshandles}{\loss_{\text{handle}}}
\newcommand{\losslayer}{\loss_{\text{Reg}}}
\newcommand{\weightloss}{\lambda}
\newcommand{\weightarap}{\weightloss_{\text{elastic}}}
\newcommand{\weightlayer}{\weightloss_\text{Reg}}
\newcommand{\weighthandles}{\weightloss_\text{handle}}
\newcommand{\energyarap}{E}
\newcommand{\bdry}{\mathbf{b}}
\newcommand{\our}{TutteNet}
\newcommand{\jac}[2]{D_{#2}#1}
\newcommand{\sregion}{\mathcal{P}}

\newcommand{\z}{\bm{z}}

\newcommand{\norm}[1]{\left\| #1\right\|}
\newcommand{\abs}[1]{\left\vert#1\right\vert}

\newcommand{\set}[1]{\left\{#1\right\}}
\newcommand{\parr}[1]{\left (#1\right )}
\newcommand{\brac}[1]{\left [#1\right ]}

\newcommand{\real}{\mathbb{R}}
\newcommand{\mats}[2]{\real^{#1\times#2}}

\newcommand{\V}{\mathbf{V}}
\newcommand{\U}{\mathbf{U}}
\newcommand{\T}{\mathbf{T}}
\newcommand{\vertex}{\mathbf{v}}
\newcommand{\uertex}{\mathbf{u}}
\newcommand{\tri}{\mathbf{t}}

\newcommand{\rthree}{\real^3}
\newcommand{\rtwo}{\real^2}
\newcommand{\bij}[2]{#1\leftrightarrow#2}

\newcommand{\rotation}{\mathbf{R}}
\newcommand{\p}{\mathbf{p}}
\newcommand{\q}{\mathbf{q}}

\definecolor{cvprblue}{rgb}{0.21,0.49,0.74}
\usepackage[pagebackref,breaklinks,colorlinks,citecolor=cvprblue]{hyperref}
\usepackage{amsmath}
\usepackage{bm}
\usepackage{overpic}
\usepackage{adjustbox}
\usepackage{amssymb}
\usepackage{pifont}
\usepackage{multirow}
\usepackage{multicol}
\usepackage{makecell}
\usepackage[bottom]{footmisc}
\def\paperID{7670} 
\def\confName{CVPR}
\def\confYear{2024}

\title{\vspace*{-20pt} TutteNet: Injective 3D Deformations by Composition of 2D Mesh Deformations}

\author{Bo Sun\\
UT Austin \\
{\tt\small bosun@cs.utexas.edu}
\and
Thibault Groueix \\
Adobe Research \\
{\tt\small groueix@adobe.com}
\and
Chen Song \\
UT Austin \\
{\tt\small song@cs.utexas.edu}
\and
Qixing Huang  \\
UT Austin \\
{\tt\small huangqx@cs.utexas.edu }
\and
Noam Aigerman \\
University of Montreal \\
{\tt\small noam.aigerman@umontreal.ca}
}

\begin{document}

\definecolor{darkgreen}{RGB}{0,110,0}
\definecolor{darkred}{RGB}{170,0,0}
\def\greencheckmark{\textcolor{darkgreen}{\checkmark}}
\def\redxmark{\textcolor{darkred}{\text{\ding{55}}}} 


\twocolumn[{
\maketitle
\vspace*{-10mm}
\begin{center}

\includegraphics[width=1.0\linewidth]{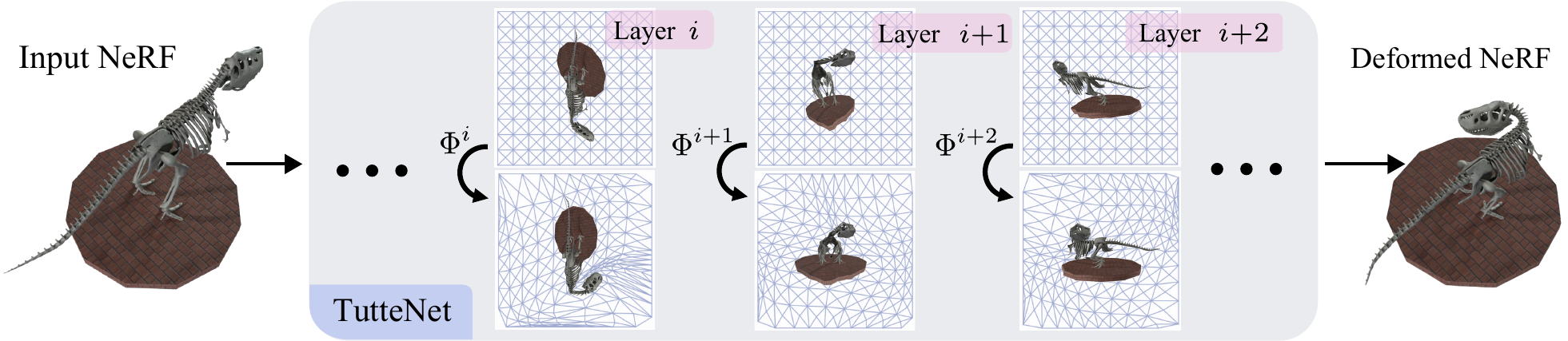}
\end{center}
\vspace*{-3mm} 
\captionof{figure}{
 Elastically deforming a NeRF~\cite{nerf} based on user-designated positioning of the head (turned) tail (bent) and body (lowered), and optimizing the degrees of freedom of \our{} to minimize the elastic energy of the deformation. \our{} guarantees an injective (1-to-1) deformation of the ambient 3D space surrounding the T-Rex, ensuring  the NeRF is rendered correctly without artifacts by enabling ``pulling back'' points and view directions from deformed space. Each layer within \our{} views the T-Rex over a different 2D plane (in this case, alternating between the 3 main axes in a tri-plane manner). In each layer, the T-Rex is enveloped with a regular 2D mesh of the unit square (top row). The 2D mesh is deformed using the layer's optimizeable parameters which define a Tutte's embedding~\cite{tutte1963draw,floater2003one} (bottom row). This defines an injective 2D piecewise-linear map, which can be applied to the 3D T-Rex, without modifying the normal direction to the plane, resulting in an injective 3D deformation $\Phi^i$. Composition of these layers yields the final expressive 3D injective deformation.
}
\vspace{3mm} 
\label{fig:teaser}
}]
\begin{abstract}
\vspace{-0.1in} 
This work proposes a novel representation of injective deformations of 3D space, which overcomes existing limitations of injective methods: inaccuracy, lack of robustness, and incompatibility with general learning and optimization frameworks. The core idea is to reduce the problem to a ``deep'' composition of multiple 2D mesh-based piecewise-linear maps.  Namely, we build differentiable layers that produce mesh deformations through Tutte's embedding (guaranteed to be injective in 2D), and compose these layers over different planes to create complex 3D injective deformations of the 3D volume. We show our method provides the ability to efficiently and accurately optimize and learn complex deformations, outperforming other injective approaches. As a main application, we  produce complex and artifact-free NeRF and SDF deformations. Our code and data are available at  \href{https://gitbosun.github.io/TutteNet}{https://gitbosun.github.io/TutteNet/}.
\end{abstract}  
\setlength{\parskip}{0pt}
\vspace{-15pt}
\section{Introduction}
\label{sec:intro}
This work concerns computation and learning of 3D deformations.
As the most immediate mode of manipulation and interaction with 3D shapes, deformations play a crucial role in various fields such as vision~\cite{wang2023tracking}, medical imaging~\cite{lv2022joint}, 3D registration~\cite{deng2022survey},  and graphics~\cite{skinningcourse:2014}. 
In many real-world applications, it is crucial that the deformation does not create any self-overlaps, i.e., is \emph{injective} (a 1-to-1 mapping). A key motivating example for this work is the deformation of  Neural Radiance Fields (NeRFs)~\cite{nerf}: when deforming NeRFs, lack of injectivity can easily cause severe rendering artifacts due to intersecting ``deformed'' rays during the ray tracing process, see Figure~\ref{Figure:nerf_cage}. 

Unfortunately, current approaches do not provide an injective deformation method that is both sufficiently expressive and robust, nor do they lend themselves to practical optimization and learning:

-- On one hand, within geometry processing and graphics,  3D deformations are heavily-researched through \emph{triangular/tetrahedral mesh} deformations, i.e., modifying the position of each vertex of the mesh. Mesh deformations provide a finite set of meaningful geometric degrees of freedom leading to stable, quick,  and straightforward computation, as well as access to geometric quantities such as the deformation gradients (\textit{Jacobians}), critical in most mesh-deformation approaches~\cite{xu2009gradient}. However, 3D mesh deformations cannot be \emph{learned} while ensuring injectivity, nor are directly applicable when an explicit triangulation of the shape is not given. 

-- On the other hand, the ML community has heavily-researched \emph{functional} representations of injective maps via neural networks, such as normalizing flows~\cite{realnvp} and solutions to ODEs~\cite{chen2018neural}. These methods were mainly designed for high-dimensional mappings, e.g., for generative tasks~\cite{grathwohl2018ffjord}, but have recently been successfully adapted to injective 3D deformations~\cite{wang2023tracking,huang2020meshode,jiang2020shapeflow}. The functional representation they provide is not geometric but rather embedded abstractly within the network's weights, often resulting in slow, cumbersome, and unstable computation, which can lead to less accurate predictions or practical inaccessibility of critical geometric quantities such as the aforementioned deformation Jacobians.

In this work, we aim to resolve these issues and gain the benefits of both worlds: we propose a novel computational representation for 3D injective deformations, which \textit{combines} the geometric representation of mesh-based deformations with the standard deep-learning approach of functional composition. 

Our core observation is that sequentially composing multiple 2D mesh deformations, over different 3D planes, achieves two critical goals simultaneously: 1) similarly to other ``deep'' representations, compositionality leads to an expressive representation, able to capture complex 3D deformations accurately, while simultaneously using mesh deformations for its layers, providing virtues such as numerical stability and accuracy; 2) while injectivity is not directly tractable for 3D mesh deformations, it is in 2D. Hence, by reducing each deformation ``layer'' to 2D, we can leverage recent observations for 2D injective mesh deformations~\cite{escher}, which show how 2D Tutte embeddings~\cite{tutte1963draw} can yield a differentiable parameterization of \emph{all} injective mesh deformations into a convex domain, enabling unconstrained learning and  optimization. Composing multiple 2D injective deformations from different viewpoint defines a family of injective volumetric 3D deformations. 

\begin{table}[t]
\centering
\setlength{\tabcolsep}{2.5pt}
\scalebox{0.82}{
\begin{tabular}{lccccc cccc}
\toprule  
& \multirow{2}{*}{Learnable} & \multirow{2}{*}{\makecell[b]{Analytical \\ inverse}} & \multirow{2}{*}{\makecell[b]{Fast Det. \\ Jacobian}}  & \multirow{2}{*}{\makecell[b]{Fast full \\ Jacobian}}  & \multirow{2}{*}{\makecell[b]{Robustness}}\\
\\
\midrule  %

i-ResNet~\cite{iresnet} & $\greencheckmark$ &  $\redxmark$ &  $\redxmark$  &  $\redxmark$ & $\redxmark$ \\ 
RealNVP~\cite{realnvp} & $\greencheckmark$ &  $\greencheckmark$ & $\greencheckmark$ & $\redxmark$ & $\redxmark$ \\
NeuralODE~\cite{chen2018neural} & $\greencheckmark$ & $\greencheckmark$ &  $\greencheckmark$ &  $\greencheckmark$  & $\redxmark$\\
\midrule  %

Ours & $\greencheckmark$  & $\greencheckmark$ & $\greencheckmark$ & $\greencheckmark$ & $\greencheckmark$ \\
\bottomrule
\end{tabular}
}
\caption{{\bf Properties of injective deformation methods.} Beyond superiority in accuracy and efficiency, our method holds unique properties compared to other injective methods, see Sec.~\ref{section:method:properties}.}
\label{tab:properties}
\vspace{-15pt} 
\end{table}
We show through experiments that our method can be used both for accurately \textit{learning} injective deformations (e.g., learning to repose a given human model to arbitrary poses), as well as \textit{optimizing} volumetric deformations in tasks in which injectivity is critical, e.g., elastically deforming a NeRF with respect to user interactions. Through comparisons, we show that our method significantly outperforms other injective deformation techniques. Furthermore, through comparisons to previous (non-injective) NeRF-deformation techniques, we both exhibit the importance of injectivity, as well as show that in many cases our method is still more expressive than those competing techniques, in spite of them facing a less-constrained problem. 

\section{Related Work}
\label{sec:related_work}

\paragraph{Invertible neural networks.} Injective maps are critical for generative modeling, in order to map between distributions. This has fueled the development of families of invertible neural functions. Normalizing Flows~\cite{dinh2014nice, germain2015made, salimans2015markov, maf, iaf, oord2016wavenet, realnvp, rezende2015variational, trippe2018conditional} are highly prominent, with RealNVP~\cite{realnvp} applied to 3D volume deformations, e.g., long-range optical flow~\cite{wang2023tracking} or for learning 3D deformations~\cite{Lei2022CaDeX,Paschalidou2021CVPR}. They work by defining an injective transformation of a subset of the spatial coordinates at each block, which is ideal for a high-dimensional settings but loses expressivity when the subsets must lie in 1D/2D. Continuous flows through Neural solutions to Ordinary Differential Equation (NeuralODE)~\cite{chen2018neural} have also been successfully applied to 3D deformations, e.g., for shape autoencoding~\cite{gupta2020neural}, dynamic mesh reconstruction~\cite{OccupancyFlow},  for point cloud generation~\cite{yang2019pointflow} and  asset deformation~\cite{huang2020meshode, jiang2020shapeflow}. Finally, i-ResNet~\cite{iresnet} achieves invertibility of a ResNet by enforcing Lipschitz bounds, with~\cite{yang2021geometryprocessingneuralfields} using this formulation for 3D deformations. We empirically evaluate and compare to these methods in Section~\ref{sec:exp:learn}. In Table~\ref{tab:properties}, we compare desirable properties for 3D invertible deformations.  \our{} can be considered as a variant of a normalizing flow, however is an explicit representation through composition of geometrically-expressive 2D mesh deformations, designed specifically for geometric 3D deformations, without the inclusion of a neural network in the deformation process.

\vspace{-4mm} 
\paragraph{Injective deformations of meshes in geometry processing.} 
\begin{figure*}[t]
\centering
\includegraphics[width=1\textwidth]{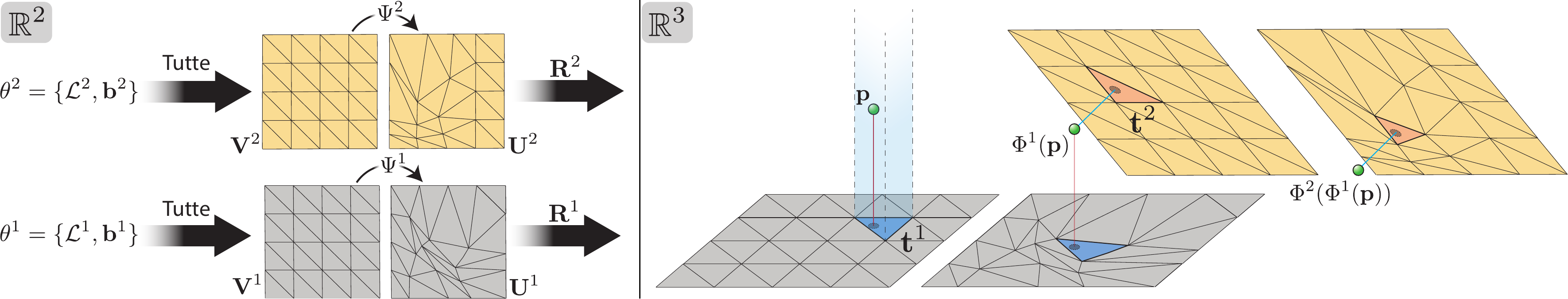}
\caption{\textbf{Our representation of injective 3D deformations}, visualized for the process of mapping a given point $\p$ inside the volume, for two-layer \our{}, $i\in\set{1,2}$. Left: the (learnable) parameters $\theta^i$, consisting of the mesh-Laplacian $\lap^i$ and the boundary conditions $\bdry^i$, define a 2D deformation $\planlayer^i$ of the square mesh $\mesh$, through Tutte's embedding~\cite{tutte1963draw}. $\planlayer^i$ is embedded in 3D to the local coordinates $\rotation^i$ to define a 3D deformation, $\layer^i$. Right: given a point $\p$, it is projected to the local coordinates of $\layer^1$, landing on triangle $\tri^1$.  $\layer^1$  defines an affine map over the infinite prism of $\tri^1$ (represented in blue with dotted lines), mapping $\p$ to $\layer^1(\p)$. The resulting $\layer^1\parr{\p}$ is projected onto the local coordinates of $\planlayer^2$, landing on triangle $\tri^2$, from which it is mapped by the affine map $\layer^2$ defined over the infinite prism of $\tri^2$.   }
\label{fig:diagram}
\vspace{-0.0in}
\end{figure*}
3D injective deformations have been extensively researched in geometry processing, mainly for piecewise linear maps on triangle meshes. Local and global injectivity can be achieved through energies that encourage or enforce it~\cite{schuller2013locally,smith2015bijective, slim,du2021optimizing,garanzha2021foldover} but they cannot guarantee injectivity when additional objectives are added, or in learning settings. Injectivity can be achieved via convex constraints~\cite{aigerman2013,kovalsky2014controlling}, tailor-made solvers~\cite{li2020incremental}, or discrete modifications of the triangulation to preserve or recover injectivity~\cite{simplexassembly,jiang2017simplicial,muller2015air} that cannot be applied in a learning setting or without a well-defined triangle mesh. Other methods use non-discrete representations~\cite{campen2016bijective} that cannot be predicted or optimized. We use layers of 2D injective deformations define via Tutte's embedding~\cite{tutte1963draw,floater2003one}, and control each  layer by the method proposed in~\cite{escher}, of optimizing the mesh Laplacian and boundary conditions. 

\vspace{-0.12in} 
\paragraph{NeRF Deformation.}
Several works use 3D volume deformations defined by MLPs that input/output spatial coordinates as a means to achieve various applications for NeRF, e.g.,  dynamic scenes~\cite{gao2021dynamicview, wu20234dgaussians, park2021hypernerf,park2021nerfies, tretschk2021nonrigid}, stylization~\cite{xu2023desrf}, and controlling trajectories~\cite{editingnerf}. Other works focus on providing NeRF deformation tools for end users, usually via a proxy geometry that controls the volume deformation, e.g., using a mesh scaffold and transfers a user-defined mesh deformation to a volume deformation Nerf-Editing~\cite{nerfediting},  or through enveloping cages~\cite{cagenerf, deformingRFCages, NeRFshop23cage, li2023interactivecage}. Others \emph{bake} the NeRFs  into a more deformation-friendly representation, such as meshes~\cite{neumesh} or point clouds ~\cite{liang2022spidr,neuraleditor}.
None of these approaches is injective nor can be applied to a NeRF without a preprocessing step of fitting the proxy to the NeRF, making learning and optimization less straightforward. We compare with \cite{nerfediting, deformingRFCages, liang2022spidr} and demonstrate the importance of injectivity. Many other learning techniques exist for deforming shapes that are not NeRFs, e.g., by predicting per-points offsets via coordinate-based MLPs~\cite{atlasnet, 3DCODED, yang2018foldingnet, fan2017pointsetgen, tancik2021learned, Park_2019_CVPR, chen2019learning,mescheder2019occupancy, deng2021deformed},  Jacobians~\cite{aigerman2022neural},  rigs~\cite{holden, li2021learning, AnimSkelVolNet, RigNet, neuroskinning,neuralcages}, or point handles~\cite{jakab2021keypointdeformer,liu2021deepmetahandles}. To avoid self-intersection, they often rely on regularizing the Jacobian~\cite{bednarik2020shape,arap, huang2021arapreg} or the Laplacian~\cite{kanazawa2018learning}.


\section{Method}
\label{method}
We now describe our expressive representation of 3D injective functions through composition of 2D injective mesh deformations (see Figure~\ref{fig:diagram} for visualization of the full pipeline). We begin by setting some necessary preliminaries regarding piecewise-linear maps and Tutte embeddings (Section~\ref{section:method:pre}), then describe our representation (Section~\ref{section:method:representation}) and conclude with analyzing its core properties (Section~\ref{section:method:properties}).

 \subsection{Preliminaries} 
 \label{section:method:pre}
 \paragraph{Piecewise-linear maps.} We assume to have a 2D triangular mesh $\mesh$ with triangles $\T$ and vertices $\V$ embedded in $\rtwo$. $\mesh$ can be any disk-topology mesh - in all experiments, we use the unit square, $\Omega=\brac{-1,1}^2$, and triangulate it with a regular triangulation of same-size isosceles triangles. We consider 2D piecewise-linear maps $\planlayer:\mesh\to\rtwo$ of this mesh, meaning the map is affine over each triangle $\tri\in\T$,  
 \begin{equation}
 \label{eq:2jac}
     \planlayer\vert_\tri\parr{\p} \equiv A_\tri \p+\delta_\tri, 
 \end{equation}
 for some $A_\tri\in\mats{2}{2},\delta_\tri\in\real^2$. (Note that this map can map \emph{any} point $\p\in\Omega$ and not just the vertices of a mesh).  The gradient of a map at point $\p$, denoted $D_\p\Psi$, is called the \emph{Jacobian}. For piecewise-linear maps, the Jacobian is constant over each triangle $\tri$ and is exactly the linear transformation $D_\tri\Psi=A_\tri$. To define a continuous piecewise linear map $\planlayer$, it suffices to define deformed vertex positions $\U=\set{\uertex_i}_{i=0}^{\abs{\V}}$, assigning position $\uertex_i\in\rtwo$ to each vertex $\vertex_i\in\V$, and define the map via $\planlayer\parr{\vertex_i} = \uertex_i$.
 Given $\uertex_i$ the Jacobian can be obtained by solving the linear equation \begin{equation}
 \label{eq:map_from_vert}
     A_\tri \vertex_i  + \delta_\tri= \uertex_i,\ i\in\tri
 \end{equation} w.r.t $A_\tri$ - the resulting small $6\times6$ linear equations can be inverted \emph{once} at initialization of training/optimization.

\paragraph{Tutte's embedding} is a method for computing injective 2D mesh mappings~\cite{tutte1963draw,floater2003one}, for meshes with disk topology (i.e., having one loop of boundary vertices). Given a mesh-Laplacian matrix $\lap$, defined by assigning some positive scalar $\lap_{ij}\in\real^+$ to each edge $\parr{i,j}$ of the mesh $\mesh$, along with a sequence of 2D points $\bdry_1,...,\bdry_k\in\rtwo$ that lie on a convex polygon, Tutte's embedding computes  deformed vertex positions $\U = \set{\uertex_i}_{i=0}^{\abs{\V}}$ by solving the sparse linear system defined via:
\begin{equation}
\label{eq:tutte}
\begin{split}
    \sum_{j}\lap_{ij}&\parr{\uertex_j-\uertex_i} = 0  \text{ for each interior vertex $\vertex_i$}\\
    &\uertex_i = \bdry_i  \text{ for each boundary vertex $\vertex_i$. }
\end{split}
\end{equation}
While Tutte's embedding is guaranteed to be injective in 2D, it is unfortunately well-known to not hold in 3D (see, e.g., ~\cite{campen2016bijective}) hence extensions to 3D do not exist. 

\subsection{3D injections through 2D mesh deformations} 
\label{section:method:representation}
We wish to devise an optimizable family of injective deformations $\map_\theta$ of 3D volumetric space, which leverages mesh deformations. Since no simple parameterization of injective 3D mesh deformations is known, the key idea of TutteNet is instead to define the 3D deformation through the composition of 2D injective mesh deformations (see Figure~\ref{fig:diagram}).

\vspace{-0.1in} 
\paragraph{Prismatic layers.}Postponing the discussion on how to compute 2D injective mesh deformations, assume for now that we have one such injective 2D mesh deformation, ${\planlayer:\bij{\mesh}{\rtwo}}$. Our basic building block constituting one ``layer'' in our architecture, is a type of map we dub a \emph{prismatic} map, meaning it is a lifting of the 2D mesh deformation $\planlayer$ into a 3D piecewise-linear map that operates over some plane and preserves the normal direction. Specifically, let $\p\in\rthree$ be a 3D point, and define
\begin{equation}
\liftlayer\parr{\p_x,\p_y,\p_z} \triangleq\planlayer\parr{\p_x,\p_y},\p_z, 
\end{equation}
i.e., $\liftlayer$ acts on the $xy$ coordinates of each point and preserves the $z$ coordinate. By rotating the coordinate system by a 3D rotation  $\rotation\in SO\parr{3}$ we can apply the deformation on any desired plane instead of on the main axes:

\begin{equation}
    \layer\parr{\p} \triangleq \rotation\liftlayer\parr{\rotation^T\p}.
\end{equation}
The process of mapping through $\layer$ is summarized in Algorithm~\ref{alg:point_map}. Finally, composing multiple  $\layer^i$ (defined over different planes $\rotation^i$ and with different $\planlayer^i$) yields the final 3D deformation $\map$,
\begin{algorithm}[b]
    Rotate point to local coordinate frame: $\q = \rotation^T\p$

    Keep only the $xy$ coordinates: $\Tilde{\q} = \parr{\q_x,\q_y}$
    
    Find triangle $\tri$ that contains $\tilde{\q}$

    Map through $\Psi$: $\tilde{\textbf{r}}  = A_\tri \tilde{\q} + \delta_\tri$

    Concatenate the $z$ coordinate back: $\mathbf{r} = \parr{\Tilde{\mathbf{r}}_x,\Tilde{\mathbf{r}}_y,\q_z}$

    Rotate back to global coordinates: $\layer\parr{\p} = \rotation \mathbf{r}$

    \caption{Prismatic map  $\layer\parr{\p}$}
    \label{alg:point_map}
\end{algorithm}

\begin{equation}
\label{eq:map}
    \map = \layer^k\circ\layer^{k-1} ... \circ\layer^0.
\end{equation}

Lastly, we need to parameterize the injective 2D mesh deformation space $\planlayer: \mesh\to\rtwo$. Our reduction of the 3D injective problem to 2D sub-problems enables us to take advantage of recent advances in 2D injective mesh deformations~\cite{escher}. Originally designed for generative 2D techniques, \cite{escher} provides a parameterization of all 2D injective deformations of a given mesh, via Tutte's embedding~\cite{tutte1963draw,floater2003one,aigerman2015}. 

Following~\cite{escher}, we use the entries of the Laplacian $\lap_{ij}$ and the boundary conditions $\bdry_i$ (defined in Section~\ref{section:method:pre}) as optimizable/learnable parameters, which in turn produce the deformed vertices $\U$ of the 2D mesh, through Tutte's embedding, by solving the linear system of Eq.~\eqref{eq:tutte} w.r.t. $\lap,\bdry$.  Following the proof in~\cite{escher}, this  covers \emph{all} possible piecewise-linear maps of $\mesh$ into the convex polygon $\bdry$.  

To ensure that $\set{\bdry_i}$ form a convex polygon, we parameterize them via positive angle increments $\alpha_i>0,\sum \alpha_i=2\pi$, and define the angle $\beta_j = \sum_{i=1}^j{\alpha_i}$. $\bdry_i$ is then the intersection of the line at angle $\beta_i$ with the unit square. Hence, $\bdry$ is a function of $\alpha$.  To keep all parameters positive and bounded, we add a sigmoid and scaling function on those parameters before feeding them to the Tutte layer,
 $x'=\text{sigmoid}\parr{x}(1-2\epsilon)+\epsilon$,
 with $\epsilon=0.2$ for $\lap$ and $\epsilon=0.1$ for $\bdry$.

 The parameters of a  prismatic map $\layer^i$ are thus $\theta^i=\parr{\lap^i,\bdry^i}$, and the local coordinate system $\set{\rotation^i}$. The final 3D injective piecewise-linear map $\map_\theta$ is thus parameterized by $\theta=\set{\theta^i,\rotation^i}_{i=0}^n$. We summarize the computation of $\map_\theta$ in Algorithm~\ref{alg:map_compute}. $\theta$ (and possibly $\set{\rotation^i}$) can be directly optimized with respect to an objective (Section~\ref{sec:exp:arap}), or otherwise predicted by a neural network (Section~\ref{sec:exp:learn}) -
 in both cases, we run Algorithm~\ref{alg:map_compute} at each iteration, compute the loss and back-propagate gradients back to $\theta$, as all steps in the algorithm are differentiable.

\vspace{-5mm}

\paragraph{Layer regularization.}
To better-condition our architecture, we can regularize its layers, ensuring each layer's Jacobian's distortion is low. We define an elastic energy measuring the Cauchy-Green strain tensor's deviation from the identity matrix $I$ at a given point $\p$ for a given map $g$, 
\begin{equation}
\label{eq:arap_energy}
    \energyarap_{g}\parr{\p} = \norm{\jac{g}{\p}^T\jac{g}{\p}-I}^2,
\end{equation}
where $\jac{g}{\p}$ is the map's Jacobian (defined in Section~\ref{section:method:pre}).
\begin{algorithm}[b]

     \caption{Computation of $\map_\theta$ from $\theta$}
        \label{alg:map_compute}
     \For{each deformation layer $i$}{
        Compute $\U^i$ via Tutte's embedding, by solving the linear system~\eqref{eq:tutte} defined by $\lap^i,\bdry^i$

        Compute $\planlayer^i$ from $\U^i$ using Eq.~\eqref{eq:2jac}, and store $A_\tri^i,\delta_\tri^i$ for each triangle $\tri$

        Define $\layer^i$ via $\planlayer^i$ and $\rotation^i$
    }
    Define $\map_\theta$ using all $\set{\Phi^i}_{i=0}^n$ via Eq.~\eqref{eq:map}
    
\end{algorithm}

\begin{figure*}[t]
\centering
\begin{overpic}[width=1\textwidth]{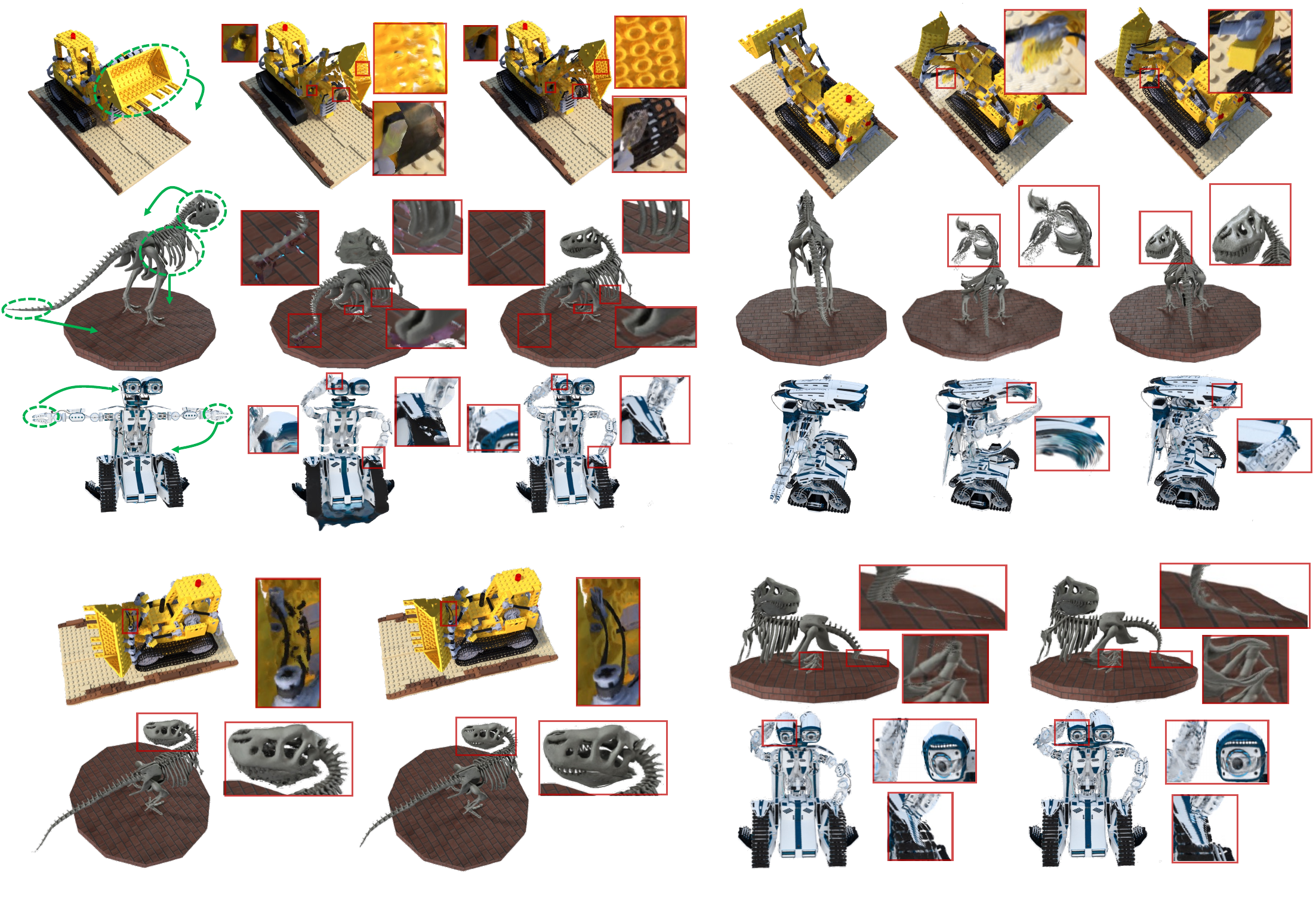}
\put(14.5, 0){\footnotesize SPIDR~\cite{liang2022spidr}}
\put(35, 0){\footnotesize TutteNet (Ours)}
\put(59, 0){\footnotesize NeuralODE~\cite{chen2018neural}}
\put(81, 0){\footnotesize TutteNet (Ours)}
\put(8.5, 27){\footnotesize View 1}
\put(21, 27){\footnotesize NeRF-Editing~\cite{nerfediting}}
\put(39, 27){\footnotesize TutteNet (Ours)}
\put(60, 27){\footnotesize View 2}
\put(70.5, 27){\footnotesize Deforming-nerf~\cite{deformingRFCages}}
\put(87.5, 27){\footnotesize TutteNet (Ours)}
\end{overpic}
\vspace{-0.1in}

\caption{{\bf Comparing  NeRF deformation methods.} We minimize the elastic deformation energy of NeRFs under user-specified constraints (left, in green) and compare the visual quality of our results with other techniques. Non-injective methods such as NeRF-Editing~\cite{nerfediting} and Deforming-NeRF~\cite{deformingRFCages} lead to non-injective deformations due to internal inversions and intersections, in turn leading to visible artifacts.  SPIDR~\cite{liang2022spidr} relies on a hybrid SDF/point cloud representation, leading to degradation in detail (T-Rex teeth) as well non-injective artifacts (tractor). 
 We additionally compare to the only other injective method that is applicable for this experiment, NeuralODE~\cite{chen2018neural} whose injectivity avoids visual artifacts, but causes \textit{geometric} artifacts such as squashing the T-Rex's tail and the robot's eye.
}
\label{Figure:nerf_cage}
\vspace{-0.1in} 
\end{figure*}

As opposed to functional representations~\cite{realnvp,huang2020meshode}, the layer's Jacobians are enumerable (one per triangle), enabling us to compute the \emph{exact} integral of $\energyarap$  (as opposed to an approximation via sampling), by summing the energy over all Jacobians of the layer:
\begin{equation}
    \losslayer \triangleq \int_{\p\in\Omega} \energyarap_{\planlayer^i}\parr{\p} \equiv \sum_{\tri\in\T}\abs{\tri}\energyarap_{\planlayer^i}\parr{\tri},
\end{equation}
where $\abs{\tri}$ is the area of triangle $\tri$. This technique could be extended in the future to, e.g., provide absolute bounds on the distortion of each layer~\cite{kovalsky2014controlling}.

\vspace{-0.01in} 
\subsection{Discussion: properties of the deformation $f_\theta$}
\label{section:method:properties}

Table~\ref{tab:properties} compares different injective approaches for 3D deformations. 
Constructing $f$ through composition of mesh deformations provides it with the following desirable properties:
\\$\bullet$ \textbf{Learnable and optimizable.}
The representation of the deformation $\map_\theta$ through the unconstrained parameters $\theta$ in turn allows simple gradient-based  learning/optimization.
\\$\bullet$ \textbf{Injective, with an immediate, explicit inverse.}
 $\map$ is guaranteed to be an injective piecewise-linear map, and  we can swap the roles of $\V^i$ and $\U^i$ to immediately get the inverse  map as well as the inverse's Jacobian.
 \\$\bullet$ \textbf{Easy and fast Jacobian computation.} 
Computing the Jacobian (see supplemental) requires $2n$ multiplications of small $3\times3$ matrices, where $n$ is the number of  layers. In comparison,  methods such as RealNVP~\cite{realnvp} are designed to have efficient access to the \textit{determinant} of the Jacobian, but require $n$ multiplications of large, dense matrices to get a full Jacobian. This is even worse when second-order optimization is needed, e.g., when the Jacobians are involved in a loss (Section~\ref{sec:exp:arap}).
\\$\bullet$ \textbf{Robust and expressive.} Our framework inherits the virtues of mesh-based deformation, e.g., numerical stability, and ability to represent elaborate deformations, with this expressivity boosted by the ability to create deep compositions of these deformations.  
Each Tutte layer relies on a single well behaved linear system, solved with a constant memory footprint and speed. In contrast, NeuralODE~\cite{chen2018neural} poses a hard-to-tune tradeoff between accuracy, speed, and memory consumption.

\begin{figure}[t]
\centering
\begin{overpic}[width=1\columnwidth]{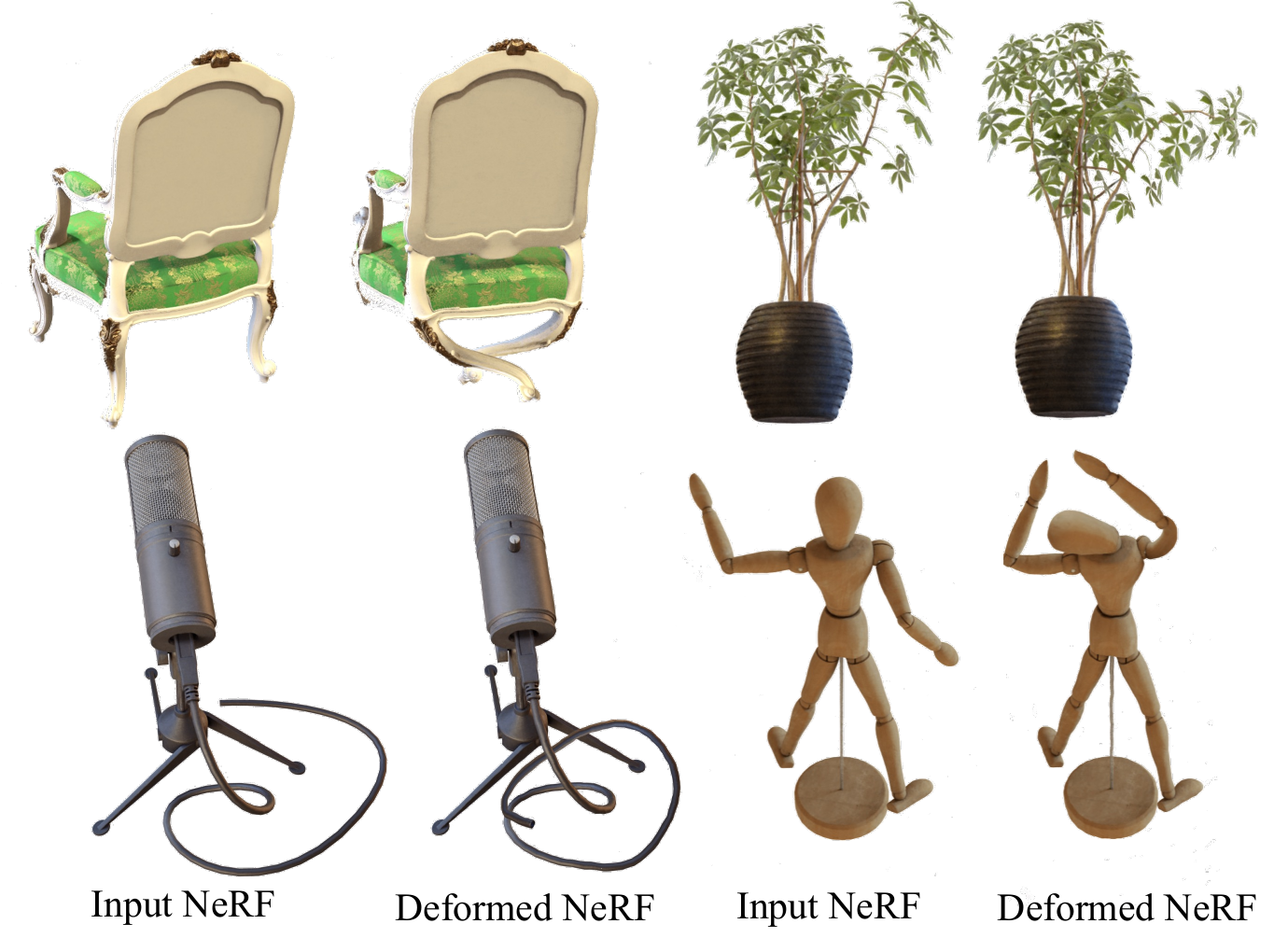}
\end{overpic}
\caption{{\bf Elastic deformations of NeRFs}. Our method guarantees injectivity and enables complex deformations, like tying a loop on the microphone's cord.}
\label{Figure:more_nerf}
\vspace{-0.1in } 
\end{figure}

\section{Experiments}
\label{sec:exp}
We evaluate the capabilities of our injective representation both on learning a space of deformations, as well as within an optimization setting. We additionally compare to several state of the art methods for deformations - both ones that are injective as well as ones that are not. In the supplementary material, we ablate the main design choices of our method \ie number of layers, resolution of each layer, as well as the orientation of the projection planes.




\subsection{Deformation of Neural Radiance Fields}
\label{sec:exp:arap}

 Neural Radiance Fields (NeRFs)~\cite{nerf} are quickly becoming one of the most popular representations for 3D scenes. Applications that use NeRFs thus require methods to manipulate and deform them, and significant research has been dedicated to NeRF deformation methods~\cite{nerfediting,NeRFshop23cage,cagenerf}. We represent a NeRF using Instant-NGP~\cite{InstantNGP},  and render it using NeRFStudio~\cite{nerfstudio}, by interfacing with their code and modifying the sampling function to go through our deformation, as we explain next.

\paragraph{Rendering the deformed NeRF.} 
 
 As discussed in previous works~\cite{deformingRFCages}, for correct rendering of a deformed NeRF, one requires both the inverse deformation as well as its Jacobian: a NeRF $N\parr{\p,\bm{r}}\rightarrow c,\sigma$ maps a point $\p$ and a view direction  $\bm{r}$ to  color $c$ and density $\sigma$, thus renderable via a ray-tracing process. Given a deformation $\map$, the point $\p$ and ray $\bm{r}$ in the deformed space correspond to the point $\p'\triangleq\map^{-1}\parr{p}$ ray $\bm{r}'\triangleq\jac{\map^{-1}}{\p}\cdot\bm{r}$ in undeformed NeRF space. Hence, we require efficient computation of $\map^{-1},\jac{\map^{-1}}{\p}$  in order to compute $N\parr{\p',\bm{r}'}$. 
 
 The only injective method, aside from ours, that supports an efficient computation of the inverse  \emph{and} its Jacobian is NeuralODE~\cite{chen2018neural,huang2020meshode,jiang2020shapeflow} - we compare to this method and show our robustness and higher accuracy. We additionally compare to non-injective methods and show the criticality of injectivity.  
 
\begin{figure}[t]
\centering
\begin{overpic}[width=1\columnwidth]{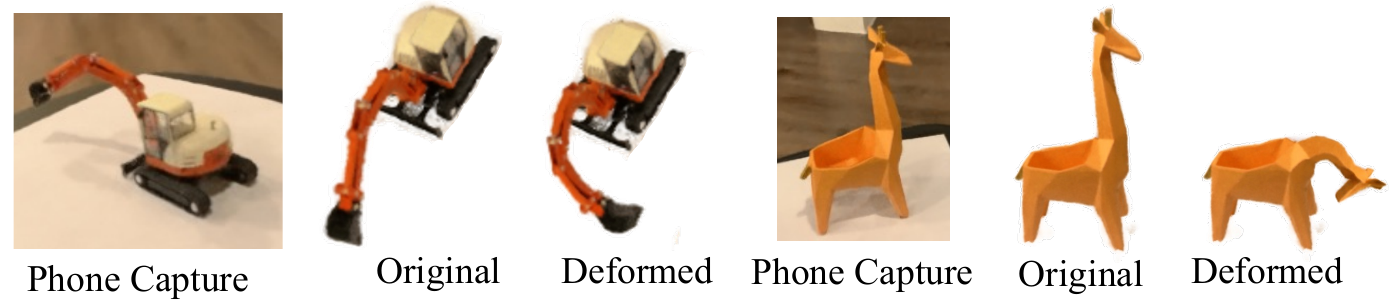}
\end{overpic}
\caption{{\bf Elastic deformations of in-the-wild NeRFs}. Our method is applicable to produce deformation of lower-quality NeRFs, captured using a phone app (Record3D). }

\label{Figure:real_nerf}
\vspace{-0.1in } 
\end{figure}

\vspace{-3pt} 

\paragraph{Elastically deforming the NeRF.} In order to deform the NeRF, we optimize the map $\map$ to satisfy the user-specified constraints in a variational as-rigid-as-possible~\cite{arap} manner,  minimizing the elastic energy, Equation~\ref{eq:arap_energy}.  

Previous methods use constructions such as proxy ``rigs'', e.g., cages~\cite{cagenerf} or point clouds~\cite{liang2022spidr}, leading to inaccuracies (e.g., when recovering the rig's geometry from an inaccurate NeRF, or when mapping between the NeRF and the rig). Our guaranteed injectivity enables deforming NeRFs directly without the tedious, brittle, proxy construction process, and we define the elastic energy of $\map_\theta$ over the density field itself:
\begin{equation}
\label{eq:lossarap}
    \lossarap = \int_{\Omega}\energyarap_{\map_\theta}\parr{\p}\sigma\parr{\p},
\end{equation}
where the integral is over  the volumetric unit cube $\Omega$,
 $\energyarap_{\map_\theta}\parr{\p}$ is defined in Equation~\eqref{eq:arap_energy}, and $\sigma$ is the NeRF's density function. The constraints are set by a user through a simple GUI,  selecting a region $\sregion$ as a ``handle'' and shifting it by a rigid motion $R$ to a new position $R\parr{\sregion}$. We enforce these constraints through an additional loss term,

\begin{equation}
    \label{eq:losshandles}
 \losshandles=\int_\sregion \norm{\map\parr{\p}-R\parr{\p}}^2.
\end{equation}
We estimate these integrals by rejection sampling on the density function $\sigma$, using a threshold of 1 on its value. Finally, we optimize the parameters $\theta$ of $\map_\theta$ w.r.t. the loss,
\begin{equation}
    \loss = \weightarap \lossarap + \weighthandles \losshandles + \weightlayer \losslayer.
\end{equation}

We show rendered NeRFs deformed by our method in  Figure ~\ref{Figure:more_nerf}, where we show our capability to support elaborate deformations, such as making a knot in the microphone chord. 
We additionally show results of elastic deformations of in-the-wild NeRFs in Figure~\ref{Figure:real_nerf}. These NeRFs were captured on an iPhone 14 with using the app Record3D. We use Nerfacto~\cite{nerfstudio} as our NeRF base model. To extract  solely the captured object,  we ignore dump points with density larger than 10, and crop points around the target objects. The deformation and rendering pipeline remains the same as described before. 
  
\vspace{-0.1in} 
\paragraph{Comparison to other NeRF-deformation approaches.}  We compare our method with three state-of-the-art NeRF deformation techniques~\cite{nerfediting, liang2022spidr, deformingRFCages} in Figure~\ref{Figure:nerf_cage}. For each method, we optimize its deformation $g$ with respect to its degrees of freedom, fitting it to the injective deformation $\map$  produced by our method.  We sample points and minimize the $L^2$ distance between the images of corresponding points, $\sum_{\p}\norm{\map\parr{\p}-g\parr{p}}$, by optimizing  the deformation's degrees of freedom with respect to this loss. We perform these tests using each method's own deformation and rendering code. 


These alternative methods focus on interactivity and speed, often losing injectivity of 3D volumetric space when fitted to strong deformations, thus resulting in artifacts. 
 Deforming-nerf~\cite{deformingRFCages} deforms the NeRF by building a cage around it and moves points by linear dependencies with the cage's vertices. This low dimensional space cannot capture the desired deformation, and without careful attention easily leads to noninjective and tangled configurations which squash the head of the T-rex and lead to rendering artifacts that blur the texture of the robot's head. Nerf-Editing~\cite{nerfediting}'s deformation of the tail of the T-rex creates an entanglement of rendering rays with the brick floor, creating ``bleeding'' artifacts in the rendering (see zoom-in). SPIDR~\cite{liang2022spidr} bakes the NeRF into a point cloud, and we used their dataset. When deformed, it can lead to a ``discrete'' version of non-injectivity, mixing points, resulting in merged teeth for the T-Rex and incorrect rendering of the Lego model.
In contrast, our deformations remain plausible and crisp, for large displacements and for diverse shapes.

\definecolor{navyblue}{RGB}{191, 209, 229} 

\begin{table}[!t]
\centering
\small %
\setlength{\tabcolsep}{2.5pt}
\begin{adjustbox}{max width=\linewidth}
\begin{tabular}{l c c  c c | c c c} 
\hline
     \multirow{2}{*}{} & \multicolumn{2}{c}{Fitting} & \multicolumn{2}{c|}{Learning} &  \multicolumn{2}{c}{Timing (sec.)} \\ 
      & Vert. $\downarrow$ & Grad. $\downarrow$ & Vert. $\downarrow$ & Grad. $\downarrow$ & Forward  & Jacobian \\ 
     \hline
     RealNVP~\cite{realnvp} & 1.7 & 12.5 & 4.21 & 35.2 & \multicolumn{1}{c}{0.006}  & 136 \\ 
     i-Resnet~\cite{iresnet}  & 17.9 & 40.2  & 13.4 &  92.3 & \multicolumn{1}{c}{\cellcolor{navyblue}0.005}  & 39 \\ 
     NeuralODE~\cite{chen2018neural} &  0.19 & \cellcolor{navyblue} {2.6} &  1.24 & 25.4 &\multicolumn{1}{c}{0.11} & 0.19 \\
     \our{} (ours) & \cellcolor{navyblue} {0.15} &  4.4  & \cellcolor{navyblue}0.16 & \cellcolor{navyblue}7.7 & 0.09 & \cellcolor{navyblue}0.02 \\  
    \hline
\end{tabular}
\end{adjustbox}
\label{tab:quantitative_comp}
\caption{{\bf Quantitative comparison of injective deformation methods.} We compare the ability of our \our{}, i-ResNet~\cite{iresnet}, RealNVP~\cite{realnvp}, and NeuralODE~\cite{chen2018neural} on the human deformation fitting and learning experiments, Section~\ref{sec:exp:learn}. We report the vertex and mesh gradient terms from Equation~\ref{eqn:vertex_jacobian_loss}, both multiplied by $10^3$. 
We report average timings on the right.}
\label{tab:quanitative_comp}
\vspace{-0.1in}
    
\end{table}

\paragraph{Comparison to NeuralODE~\cite{chen2018neural,huang2020meshode,jiang2020shapeflow}.} As discussed above, NeuralODE is the only other method that provides both inverse and Jacobians in a computationally-feasible manner (e.g., not resorting to second-order derivation of an MLP when optimizing the Jacobian-dependent energy, Equation~\eqref{eq:lossarap}). We replaced our representation with theirs and ran the experiment with exactly the same setup. Results are shown in Figure~\ref{Figure:nerf_cage}. This comparison highlights the importance of injectivity for NeRF deformation, as neither of the two methods exhibits rendering artifacts in any scenario. However, the zoom-ins reveal geometric issues: NeuralODE  completely collapses part of the T-Rex's leg, and squashes the eye and hand of the robot, due to their proximity to one another. We additionally note that NeuralODE's numerical integration sometimes leads to running out of memory or significant stalling, when run on a large set of sample points, reducing its applicability to the NeRF rendering setting.

\begin{figure}[t]
\centering
\begin{overpic}[width=1\columnwidth]{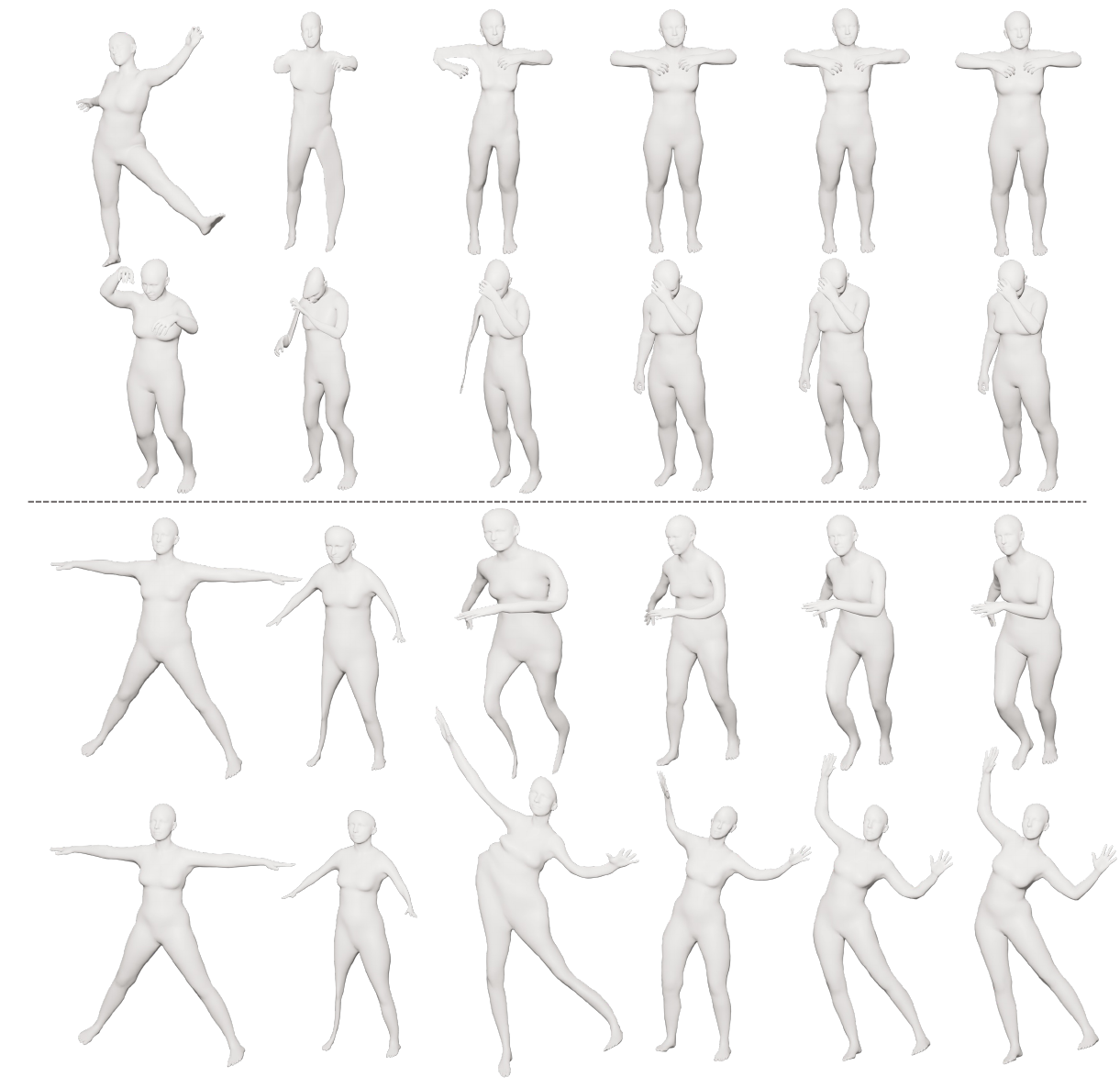}
\put(1, 17) {\rotatebox{90}{Learning}}
\put(1, 70) {\rotatebox{90}{Fitting}}
\put(10, -1) {\footnotesize Source}
\put(25, -1) {\footnotesize i-Resnet~\cite{iresnet}}
\put(42, -1) {\footnotesize NVP~\cite{realnvp}}
\put(57, -1) {\footnotesize ODE~\cite{chen2018neural}}
\put(74.5, -1) {\footnotesize Ours}
\put(89, -1) {\footnotesize GT}
\end{overpic}
\caption{{\bf Visual comparison of accuracy of injective deformation methods.} We compare the ability of our \our{}, i-ResNet~\cite{iresnet}, RealNVP~\cite{realnvp}, and NeuralODE~\cite{chen2018neural} on fitting (top) and learning (bottom) human deformations, Section~\ref{sec:exp:learn}. Our method produces highly-accurate results in the learning experiment while all others show visible artifacts. For the fitting experiment, only  NeuralODE~\cite{chen2018neural} achieves similar accuracy to ours.}

\label{Figure:fitting_comp}
\vspace{-0.1in } 
\end{figure}
\subsection{Learning Injective Deformations}
\label{sec:exp:learn}
We evaluate the applicability of \our{} in a learning setting. Here, the parameters $\theta$, which define the deformation $\map_\theta$, are predicted by a neural network - note that as opposed to a standard ``hypernetwork'' which predicts parameters of another neural network, here the neural network predicts geometrically meaningful degrees of freedom and hence we do not expect significant degradation in accuracy. To quantify and evaluate our representation's ability to accurately capture injective deformations, we require a dataset with ground truths, and hence we choose to use the highly popular SMPL~\cite{smpl} model, which can generate a dataset of human meshes with groundtruth 1-to-1 correspondences between their vertices.

\begin{figure}[t]
\centering
\begin{overpic}[width=0.48\textwidth]{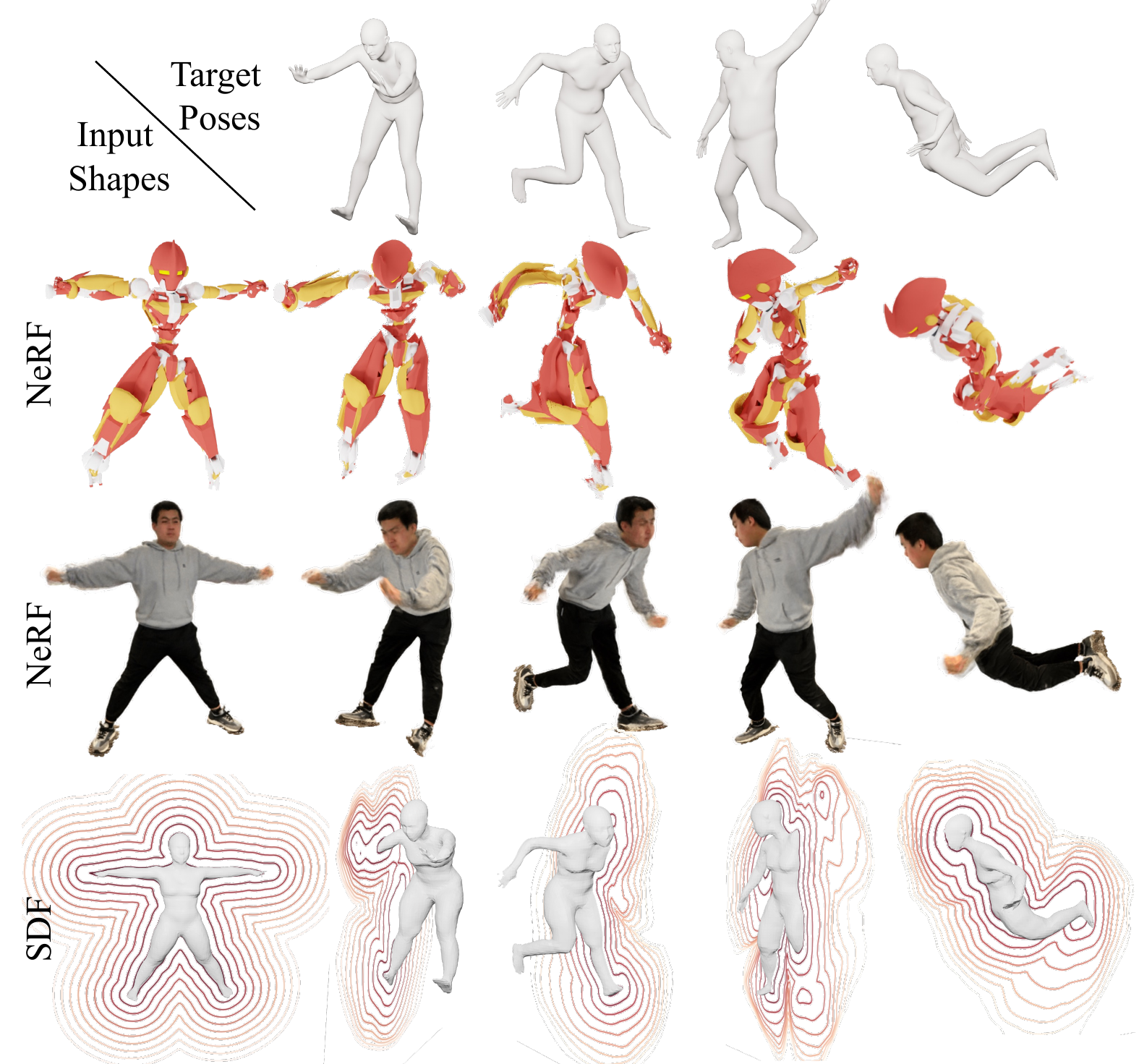}
\end{overpic}
\caption{\bo{ {\bf Deforming different neural fields using the same trained network.} The neural network from the learning experiment (Section~\ref{sec:exp:learn}),  trained to predict deformations on a dataset of human SMPL~\cite{smpl}  meshes (top row, demonstrating the desired target pose), is seamlessly applied to deform synthetic and real \textit{NeRFs}~\cite{nerf} (middle two rows), and SDFs~\cite{Park_2019_CVPR} (bottom row). }}
\label{Figure:learning_nerf}
\vspace{-0.1in}
\end{figure}

SMPL is parameterized by two sets of parameters: $P$ and $B$, dictating the human pose and body shape, resp. We generate a dataset of different body-shaped humans, each in a pair of source and target poses, $S_{B,P_s},S_{B,P_t}$. We randomly sample poses, discarding results with self-intersections. See the supp. material for full details.

Our training scheme trains the neural network to receive the source human $S_{B,P_s}$, and the target pose parameters $P_t$, and based on them predict the deformation $\map_\theta$ that deforms the source to the target $\map_\theta\parr{S_{B,P_s}}=S_{B,P_t}$. To avoid inference that relies on specific geometric structure, we encode each source human $S_{B,P_s}$ by rendering it from several viewpoints and using a visual encoder. We concatenate the output of the encoder along with the pose parameters $P_t$ into a code $\z$, which is fed into an MLP architecture that predicts the final deformation parameters $\theta$. We compute the map $\map_\theta$ via Algorithm~\ref{alg:map_compute} and use it to compute the same loss used by~\cite{aigerman2022neural} for learning mesh deformations:
{\small
\begin{equation}
\begin{aligned}
\label{eqn:vertex_jacobian_loss}
    \loss = &\norm{\map_\theta\parr{S_{B,P_s}}-S_{B,P_t}}^2+  \\
    0.1 &\norm{J\map_\theta\parr{S_{B,P_s}}-JS_{B,P_t}}^2,
\end{aligned}
\end{equation}
}
where the first term is the $L_2$ distance between the deformed human mesh's vertices and the ground-truth target vertices, and the second term is the $L_2$ distance between the deformed mesh's and the ground truth mesh's intrinsic deformation gradient (note: \emph{not} the map's Jacobians), obtained by using the source mesh's gradient operator. See supplementary for full details on training. Figure~\ref{Figure:fitting_comp}, bottom shows the predicted deformations. See more results in the supp. material. 

\

\begin{table}[t]
 
\begin{subtable}[h]{0.48\textwidth}
\centering
\begin{adjustbox}{max width=0.8\linewidth}
\begin{tabular}{l | c c  c c } 
    \hline
    Mesh Resolution & 7 & 11 & 17 & 25 \\ 
    \hline
     Fitting Vert. & 0.22 & 0.15 & 0.11 & 0.09 \\
     Fitting Grad. & 5.6 & 4.4 & 2.9 & 2.1 \\
     \hline
     Forward Time & 0.088 & 0.094 & 0.118 & 0.166 \\
     Jacobian Time & 0.02 & 0.02 & 0.02 & 0.02 \\
    \hline
\end{tabular}
\end{adjustbox}
\caption{Ablation study on the mesh resolutions. The number of layers are fixed to 24 with the tri-plane architecture. }
\label{supp:tab:ablation_reso}
\end{subtable}
\hfill
\\
\\
\begin{subtable}[h]{0.48\textwidth}
\centering
\begin{adjustbox}{max width=0.8\linewidth}
\begin{tabular}{l | c c c c } 
\hline
     Num of Layers & 6 & 12 & 24 & 36 \\ 
     \hline
     Fitting Vert. & 1.29 & 0.24 & 0.15 & 0.12 \\
     Fitting Grad. & 12.5 & 6.1 & 4.4 & 3.2 \\
     \hline
     Forward Time & 0.025 & 0.048 & 0.094 & 0.142 \\
     Jacobian Time & 0.005 & 0.011 & 0.020 & 0.029 \\
    \hline
\end{tabular}
\end{adjustbox}
\caption{Ablation study on the number of layers. Mesh resoution is fixed to $11\times11$ vertices. }
\label{supp:tab:ablation_layer}
\end{subtable}
 \caption{\textbf{Ablation studies on the Tutte mesh resolutions and number of Tutte layers}. We report the $L_2$ and Grad. errors as well as forward and Jacobian times on the fitting experiment. In the main experiments, parameters are chosen based on the trade-off of the speed and accuracy.}
 \label{supp:tab:ablation}
 \vspace{-0.1in}
\end{table}

Since the network (trained solely on meshes) produces volumetric injective deformations, it can be readily applied to other neural fields  such as NeRFs~\cite{nerf} and SDFs~\cite{Park_2019_CVPR} -  Figure~\ref{Figure:learning_nerf} shows results of applying the same network, without retraining, on: 1) synthetic NeRF created from a rendered model; 2) real in-the-wild NeRF captured by a smartphone; 3) SDF, showing the SDF isolines as well as the marching cube reconstruction (note, though, that any deformation of an SDF violates the Eikonal equation. Hence, while the deformed field represents a valid shape, it is no longer an SDF).
Although the network was only trained with respect to points \emph{on the surface} of the mesh, its injectivity ensures it produces meaningful deformations on the volume of the models. We additionally note that there are many methods that focus specifically on deforming NeRFs of humans (\eg SHERF~\cite{hu2023sherf}), while we use humans as a benchmark for comparing and measuring the accuracy of our method, as well as showing its generality: unlike these other techniques~\cite{chen2021animatable, neural-human-radiance-field, jiang2022selfrecon, kwon2021neural, peng2021neural, peng2023implicit, Xu2021HNeRFNR, Zhao_2022_CVPR, weng2022humannerf,su2021nerf},  we did not use any human-specific priors in the design of the representation, and the same exact method could be applied as-is to any other deformation dataset.


\paragraph{Comparison to other methods for learning injective deformations.} We use the learning experiment to compare our method with other key representatives of families of invertible neural representations: RealNVP~\cite{realnvp} (normalizing flows), i-ResNet~\cite{iresnet} (Lipschitz-bounded networks) and NeuralODE~\cite{chen2018neural} (continuous flow/ODE solutions) - all of which have been successfully applied to 3D tasks~\cite{wang2023tracking, Lei2022CaDeX, Paschalidou2021CVPR, gupta2020neural, OccupancyFlow, yang2019pointflow, jiang2020shapeflow}. We trained these methods exactly as we trained ours, with their provided code and with same number of model parameters, and performed hyperparameter sweeps to find the best-performing choices for each - refer to the supplementary. Quantitative results are shown in Table~\ref{tab:quanitative_comp}, and representative example deformations are visualized in Figure~\ref{Figure:fitting_comp}, bottom. Our method accurately learns the deformation space, achieving near-identical deformations and the lowest fitting error, while the other methods achieve higher errors, leading to visible artifacts. Our method also achieves significantly faster Jacobian computation.

 As an additional experiment, we measure the ability of each technique to represent \textit{one, single deformation}, by ``overfitting'' the network (without a conditional) on a pair of source/target humans.
For this evaluation, we randomly generated 200 different-bodied humans, and sampled pose source/target pairs from the AMASS dataset~\cite{amass} for each of them. 
We show quantitative results in Table~\ref{tab:quanitative_comp} and qualitative results in Figure~\ref{Figure:fitting_comp}. 

As is evident from both experiments, RealNVP~\cite{realnvp} and i-ResNet~\cite{iresnet} produce inaccurate results compared to us, both in the learning experiment and in the fitting experiment. Indeed, they are designed to perform extremely well on high-dimensional tasks, but are less successful in 3D tasks which require very high accuracy (note the shrunk parts in Figure~\ref{Figure:fitting_comp}). RealNVP~\cite{realnvp} has low-dimensional injective coupling layers similar to us, however, each of their layers is less expressive than a Tutte layer.  i-ResNet~\cite{iresnet} achieves invertibility by regularizing ResNet blocks to have a Lipschitz constant $<1$, however, this family of functions leads to reduced expressivity to fit 3D deformations. 

While NeuralODE~\cite{chen2018neural} produces significantly less accurate deformations in the learning experiment, for the fitting experiment, they achieve results comparable to ours, with results visually close to indistinguishable in Figure~\ref{Figure:fitting_comp}, and in fact, for the fitting experiment, attain a slightly lower average error than us on the mesh-gradient fitting term (``Grad.'') while we achieve a lower vertex fitting term, see Table~\ref{tab:quanitative_comp}. The degradation in NeuralODE's performance when scaling up for the learning experiment is expected: to achieve injectivity, they leverage uniqueness of ODE solutions and plot reversible trajectories of points in 3D space - this requires numerical integration, which becomes increasingly difficult as the learned functional space represented by the neural network grows more convoluted (refer to~\cite{chen2018neural} and their discussion on their Figure 3(d)). For the fitting experiment we used the default hyperparameters from ~\cite{chen2018neural}, and for the learning experiment used the ones from~\cite{jiang2020shapeflow}.

\subsection{Ablations}
\label{sec:ablation}
We ablate on the main parameters of our model. We first examine the effect of modifying the height (number of Tutte layers) and width (mesh resolution) of the TutteNet. We evaluate different choices for these two parameters on the fitting experiment from Section 4.2. We show results in Table~\ref{supp:tab:ablation} - we report the vertex and and gradient error, both multiplied by $10^3$, as well as average timings at the bottom. As expected, increasing any of these two parameters improves performance at the price of a slower computation.

Additionally, we validate the effect of the way the local coordinates $\rotation^i$ are chosen. We run the learning experiment (Section~\ref{sec:exp:learn}) with different policies for choosing $\rotation^i$. Results are shown in Table~\ref{supp:tab:ablation_orentation}. We show policies variants for  choosing $\rotation^i$: regular triplane (alternating between 3 canonical coordinate systems on the 3 main axes); cubic (alternating between 8 directions corresponding to the 8 corners of the cube); and predicting $\rotation^i$ using a neural network. The best option is to let a neural network control the local coordinates.

\begin{table}[h]
\centering
\small %
\setlength{\tabcolsep}{2.5pt}
\begin{adjustbox}{max width=\linewidth}
\begin{tabular}{l |c c  c } 
\hline
      & Triplane &  Cubic  &  Predicted \\ 
     \hline
     Learning Vert. & 0.25 & 0.21 & 0.16 \\
     Learning Grad. & 8.9 & 8.4 & 7.7 \\
    \hline
\end{tabular}
\end{adjustbox}
\caption{Ablation on choices of orientations.  We report the $L_2$ and Grad. errors in the learning experiment, where we use 24 layers with a mesh resolution $11 \times 11$. }
     \label{supp:tab:ablation_orentation}
\end{table} 


\vspace{-6mm}
\section{Conclusion}
\label{conclusion}

We have presented an expressive, numerically robust, and computationally efficient representation of 3D injective deformations that can be plugged without modification into other applications requiring injective deformation modules.  

Our method has two main limitations. First, the map evaluation cannot be done at interactive times, preventing real-time rendering and interaction for the moment.
Second, deforming one part of the space may have an effect on another part, and it is non-trivial to completely localize deformations to one part of a shape - however,
note that this limitation also holds for all the other injective deformation techniques.


We are excited by the possible uses of our framework, e.g., for long-range optical flow~\cite{wang2023tracking}, or to regularize non-rigid 3D registration~\cite{deng2022survey}. Additionally, use cases for other types of low-dimensional injective maps are highly attractive, e.g., for surface-to-surface mappings through common domains, which require an injective 2D map~\cite{neural_surface_maps}.

\vspace{3mm} 
\noindent \textbf{Acknowledgments}: This work was supported in part through the grants NSF IIS-2047677,
HDR-1934932, CCF-2019844, and IARPA WRIVA program.
{
    \small
    \bibliographystyle{ieeenat_fullname}
    \bibliography{main}
}

\end{document}


\definecolor{darkgreen}{RGB}{0,110,0}
\definecolor{darkred}{RGB}{170,0,0}
\def\greencheckmark{\textcolor{darkgreen}{\checkmark}}
\def\redxmark{\textcolor{darkred}{\text{\ding{55}}}} 

\maketitle

\section{Additional Results}

\begin{figure*}[h]
\centering
\begin{overpic}[width=1\textwidth]{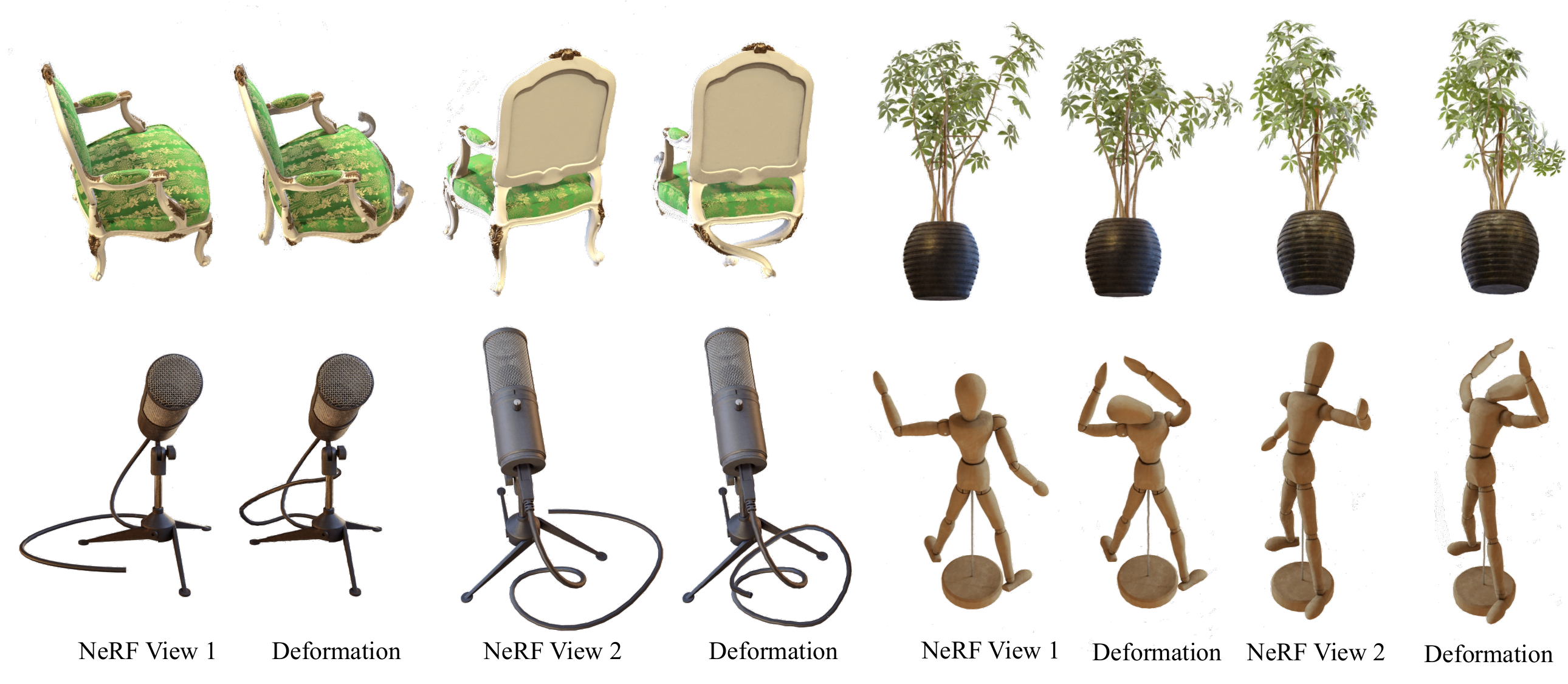}
\end{overpic}
\vspace{-0.1in}
\caption{Additional elastic deformations of various NeRFs~\cite{nerf} via our representation, as described in Section 4.1 in the main paper.}
\label{supp:Figure:more_nerf}
\end{figure*}

\begin{figure*}[h]
\centering
\begin{overpic}[width=0.8\textwidth]{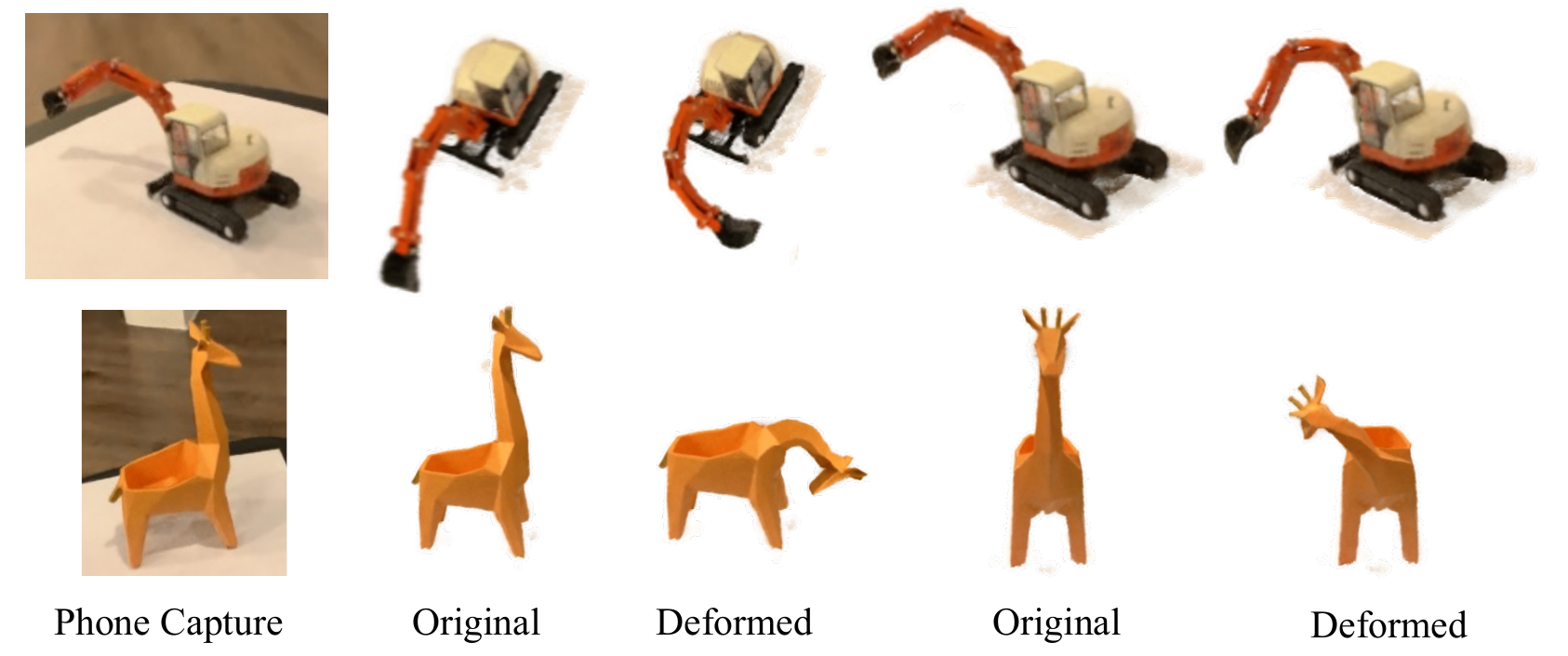}
\end{overpic}
\vspace{-0.1in}
\caption{Additional elastic deformations of phone-captured NeRFs~\cite{nerf} via our representation.}
\label{supp:Figure:real_nerf}
\end{figure*}



\subsection{Additional Results of NeRF Deformations (Section 4.1 in the main paper)}

We produced more NeRF deformations, minimizing the elastic energy of the deformation (Section 4.1 in the main paper). Results are shown in Figure~\ref{supp:Figure:more_nerf}. 
We also show more qualitative results on real NeRFs we captured with a smartphone in Figure~\ref{supp:Figure:real_nerf}. 

\subsection{Additional Results for Fitting and Learning of Human Poses (Section 4.2 in the main paper)}
\begin{figure*}[t]
\centering
\begin{overpic}[width=1\textwidth]{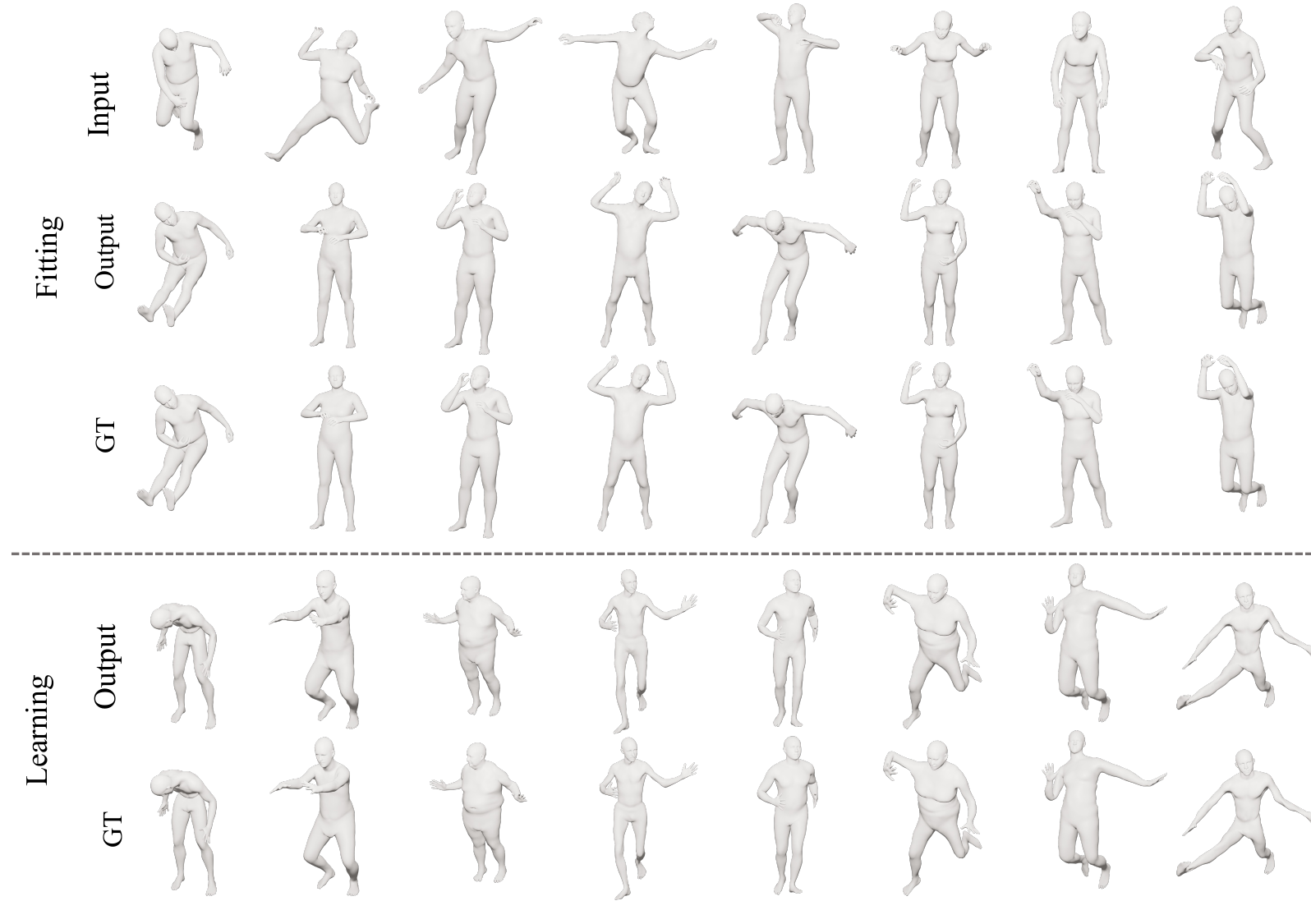}
\end{overpic}
\vspace{-0.1in}
\caption{Additional results on the learning (bottom) and fitting (top) human poses experiment (section 4.2). }
\label{supp:Figure:fitting_learning}
\end{figure*}

More results on the fitting and learning experiments can be found in Figure~\ref{supp:Figure:fitting_learning}.

\subsection{Multi-view Videos for our NeRF Deformation Results (Section 4.1 in the main paper)}
You can find the videos for 3 NeRF deformations shown in the main paper in the NeRF\_videos.zip.

\subsection{Intermediate Deformation Process }
We further show the detailed intermediate deformations of our method (see attached intermediate\_deformations.gif). Our method explicitly deforms the space by composing a sequence of 2D mesh deformations. The spatial points are deformed accordingly. By choosing different 3D orientations, we get our final 3D deformation. 

\section{Ablation Study}
\label{supp:sec:ablation}
Since our method proposes a new representation, we ablate on the main parameters of our model: number of Tutte layers, and the mesh resolution while performing the fitting experiment from Section 4.2 - this experiment directly validates the capacity of our representation to fit to deformations as we modify the ablated parameters. We show results in Table~\ref{supp:tab:ablation} - we report the vertex and mesh gradient terms (as in Table 2 in the main paper), both multiplied by $10^3$. We report average timings at the bottom. As expected, increasing any of these two parameters improves performance at the price of a slower computation.

The one additional parameter that could be ablated is the local coordinates $\rotation^i$. We ablate on them by running the learning experiment (Section 4.2 of the main paper) with different methods to choose $\rotation^i$. Table~\ref{supp:tab:ablation_orentation} shows the results. We show three variants for plane choices: regular triplane (alternating between 3 canonical coordinate systems on the 3 main axes); cubic (alternating between 8 vertex directions); and predicting $\rotation^i$ using a neural network. The best option is to let a neural network control the local coordinates.

\begin{table}[h]
	\begin{subtable}[h]{0.5\textwidth}
		\centering
		\begin{tabular}{l | c c  c c } 
\hline
     Mesh Resolutions & 7 & 11 & 17 & 25 \\ 
     \hline
     Fitting Vert. & 0.22 & 0.15 & 0.11 & 0.09 \\
     Fitting Grad. & 5.6 & 4.4 & 2.9 & 2.1 \\
     \hline
     Forward Time & 0.088 & 0.094 & 0.118 & 0.166 \\
     Jacobian Time & 0.02 & 0.02 & 0.02 & 0.02 \\
    \hline
\end{tabular}
		\caption{Ablation study on the Tutte mesh resolutions. We report the $L_2$ and Grad. errors on the fitting experiment with 8 tri-plane models.}
		\label{supp:tab:ablation_reso}
	\end{subtable}
	\hfill
	\begin{subtable}[h]{0.5\textwidth}
		\centering
		\begin{tabular}{l | c c c c } 
\hline
     Num of Layers & 6 & 12 & 24 & 36 \\ 
     \hline
     Fitting Vert. & 1.29 & 0.24 & 0.15 & 0.12 \\
     Fitting Grad. & 12.5 & 6.1 & 4.4 & 3.2 \\
     \hline
     Forward Time & 0.025 & 0.048 & 0.094 & 0.142 \\
     Jacobian Time & 0.005 & 0.011 & 0.020 & 0.029 \\
    \hline
\end{tabular}
		\caption{Ablation study on increasing the number of layers (mesh resoution is fixed to $11\times11$ vertices) We report the $L_2$ and Grad. errors on the fitting experiment (Table 2 in the main paper). }
		\label{supp:tab:ablation_layer}
	\end{subtable}
 \caption{Ablation study on the Tutte mesh resolutions and number of Tutte layers.}
 \label{supp:tab:ablation}
\end{table}

\begin{table}[h]
\centering
\small %
\setlength{\tabcolsep}{2.5pt}
\begin{adjustbox}{max width=\linewidth}
\begin{tabular}{l |c c  c } 
\hline
      & Triplane &  Cubic  &  Predicted \\ 
     \hline
     Learning Vert. & 0.25 & 0.21 & 0.16 \\
     Learning Grad. & 8.9 & 8.4 & 7.7 \\
    \hline
\end{tabular}
\end{adjustbox}
\caption{Ablation on choices of orientations in the learning experiment. In a triplane manner, we alternate orientations among $x,y,z$ axes. In a cubic manner, we chose the orientations of 8 vertices of the unit cube and alternated among those 8 directions. In a predicted manner, we use an MLP to predict orientations for all layers. In this experiment, we use 24 layers with a mesh resolution $11 \times 11$. }
     \vspace{-0.2in}
     \label{supp:tab:ablation_orentation}
\end{table}

\section{Detailed Timing Analysis}
\label{supp:sec:timing}

\begin{figure*}[t]
\centering
\begin{overpic}[width=1\textwidth]{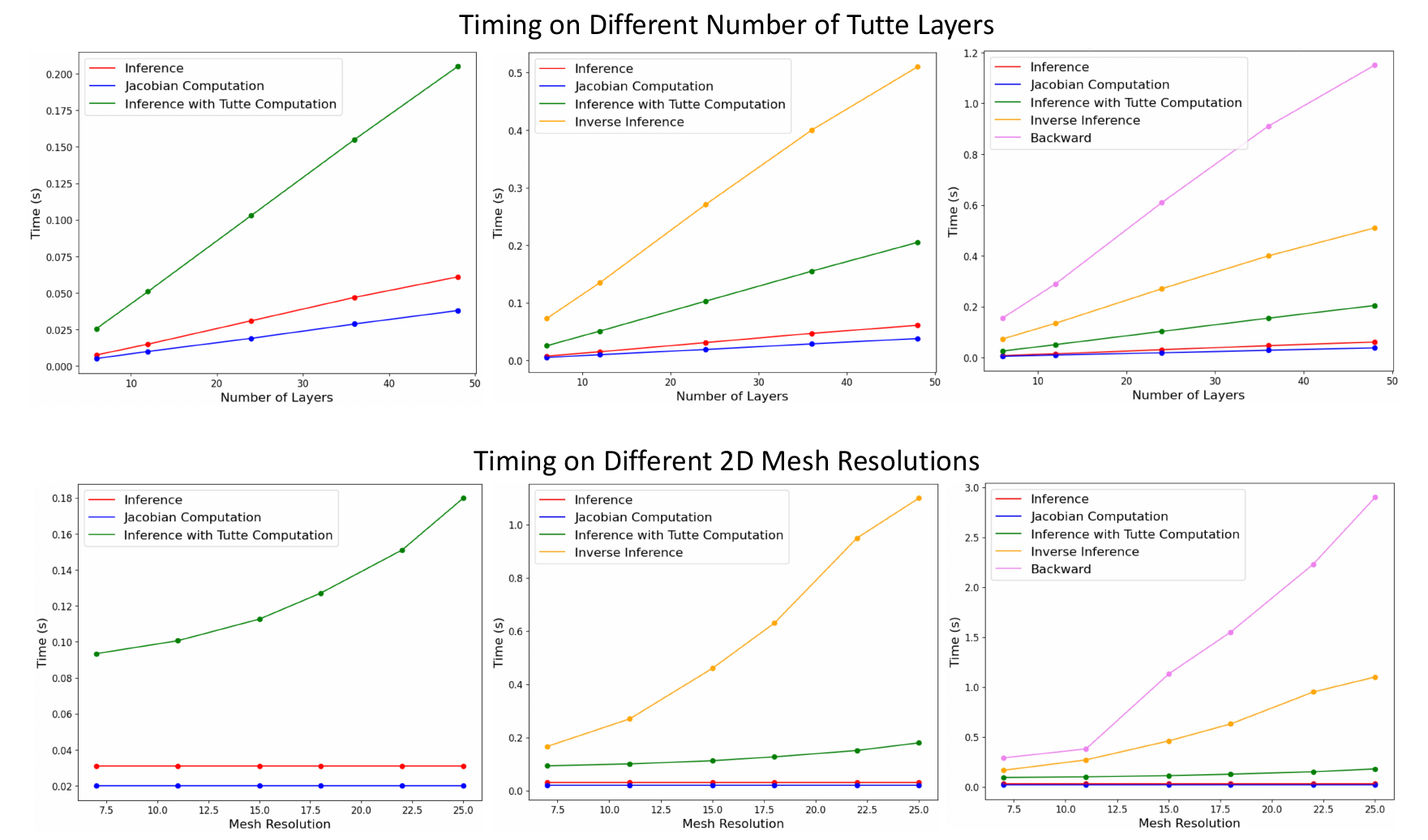}
\end{overpic}
\vspace{-0.1in}
\caption{Timing analysis on number of Tutte layers and 2D mesh resolutions. Top: timing w.r.t. number of Tutte layers. Bottom timing w.r.t. the 2D mesh's resolutions, with the fixed number of layers as 24. }
\label{supp:Figure:timing}
\end{figure*}

We show a detailed timing analysis for our method in Figure~\ref{supp:Figure:timing}. All numbers are averaged on 200 pairs in the fitting experiments (Section 4.2 in the main paper). 
With an increase of number of Tutte layers, the time increases linearly as each layer's computation takes a fixed amount of time.
The 2D mesh resolution, on the other hand, can be increased while both the Jacobian computation time and the inference time remain close to constant, as their computation per layer is not affected by mesh resolution significantly. Of course, computing the Tutte embedding (solving the linear system Eq. 3 in the main paper), the inverse time and the back-propagation time increase as the mesh becomes denser.

\section{Limitation: Non-localized Effect of the TutteNet Representation (Section 5 in the main paper)}
\begin{figure*}[t]
\centering
\begin{overpic}[width=1\textwidth]{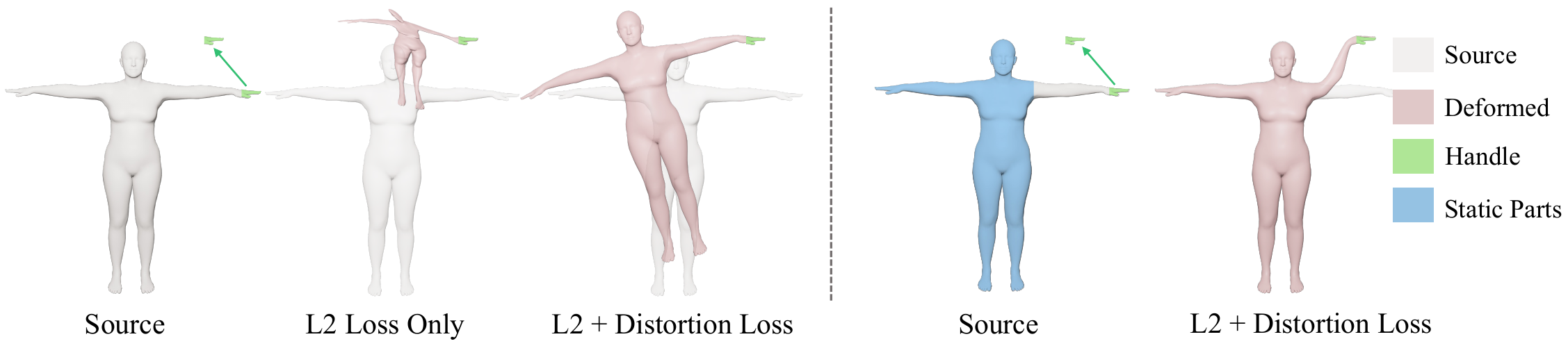}
\end{overpic}
\vspace{-0.1in}
\caption{Our deformation is not localized: we show the \textit{source} model and the constraint (green) dragging the hand to a new position. Second from left, we show the result of optimizing for the fitting of the constraint, without applying any regularization to other parts of the shape; the unconstrained part moves as the changes the TutteNet performs to fit the constraint have global effect, modifying the deformation in every part of the space. Third from the left: once we add a distortion minimization regularizer to every other part of the human, the hand goes to place, and the entire TutteNet converges into emulating the deformation which matches the constraint and minimizes the elastic energy: a global rotation. Right: The result of applying the same constraint, but while regularizing to keep the entire body of the human (blue) static, allowing only a small portion of the hand to bend. }
\label{supp:Figure:limitation}
\end{figure*}

As mentioned in Section 5 in the main paper, one limitation of our method (that all other injective methods share) is that deforming one part of space may have an effect on another part, and it is non-trivial to completely localize deformations to one part of a shape. We show an example of this issue in Figure~\ref{supp:Figure:limitation}. 
On the left, we show the source model and the constraint (green) dragging the hand to a new position. Second, from left, we show the result of optimizing for the fitting of the constraint, without applying any regularization to other parts of the shape; the unconstrained part moves as the changes the TutteNet performs to fit the constraint have global effect. Third from the left: once we add a distortion minimization regularizer to every other part of the human, the hand goes to place and the entire TutteNet converges into emulating the deformation which matches the constraint and minimizes the elastic energy: a global rotation. Right: The result of applying the same constraint, but regularizing to keep the entire body of the human (blue) static, allowing only a small potion of the hand to bend.

\section{Computation of the Jacobian of the Deformation}
The deformation's Jacobian can be computed in a quick and straightforward manner. Let $\p\in\rthree$, and define a prismatic map $\layer^i$ with respect to $\rotation^i,\planlayer^i$ as in Section 3.2. Then the Jacobian of $\layer^i$ at point $\p$ is
\begin{equation}
\label{eq:prism_jac}
    \jac{\layer^i\equiv\rotation^i}{\p} \tilde{A}_\tri {\rotation^i}^T,
\end{equation}
where $\tri$ is the triangle the point lies in (per Algorithm 1), and
   ${\tilde{A}_\tri = 
\begin{pNiceArray}{cc cc}[small]
    \Block{2-2}{A_\tri} & & 0\\ 
    & & 0 
    \\
    0 &  0 & 1
\end{pNiceArray}}$ is the 2D Jacobian of the 2D mesh deformation at point $\p$, lifted to 3D.
Finally, the Jacobian of the map $\map$ at point $\p$ can be computed by applying the chain rule to equation (6), 
\begin{equation}
\label{eq:map_jac}
    \jac{\map}{\p} = \Pi_{i=o}^n \jac{\layer^i}{\p}.
\end{equation}


\section{Training Details of the Fitting and Learning Experiments (Section 4.2 in the main paper)}
\noindent \textbf{Choices of Shape Pairs in the Shape Fitting Experiment}.
We randomly selected pairs of shapes from the AMASS training set~\cite{amass}. In order to focus our fitting experiments solely on pose changes, we aligned the shape parameters of the source shapes with those of their corresponding target shapes. However, this alignment could potentially result in self-intersections in the source shape due to the modification of its parameters. To address this, we excluded pairs with self-intersecting source shapes and retained 200 pairs for the evaluation set in our fitting experiment

\noindent \textbf{Dataset Generation of the Learning Experiment}.
\label{supp:sec:data_learning}
Our training set for the learning experiment is derived from the AMASS dataset~\cite{amass}. Rather than directly utilizing the training set provided by AMASS, which contains numerous repetitive and closely related poses, we opt to construct our dataset by randomly sampling from a Gaussian distribution based on the pose and shape distributions observed in the AMASS dataset.
To achieve this, we calculate the mean ($\mu_s, \mu_p$) and variance ($\sigma_s, \sigma_p$) for all shape and pose parameters in the AMASS training dataset. Subsequently, we sample our dataset's shape parameters with a mean of $\mu_s$ and a variance of $2\sigma_s$, while pose parameters are sampled with a mean of $\mu_p$ and a variance of $1.5\sigma_p$.
Our model is then trained on this randomly sampled dataset, and its performance is evaluated on the AMASS validation set.

\begin{figure*}[t]
\centering
\begin{overpic}[width=1\textwidth]{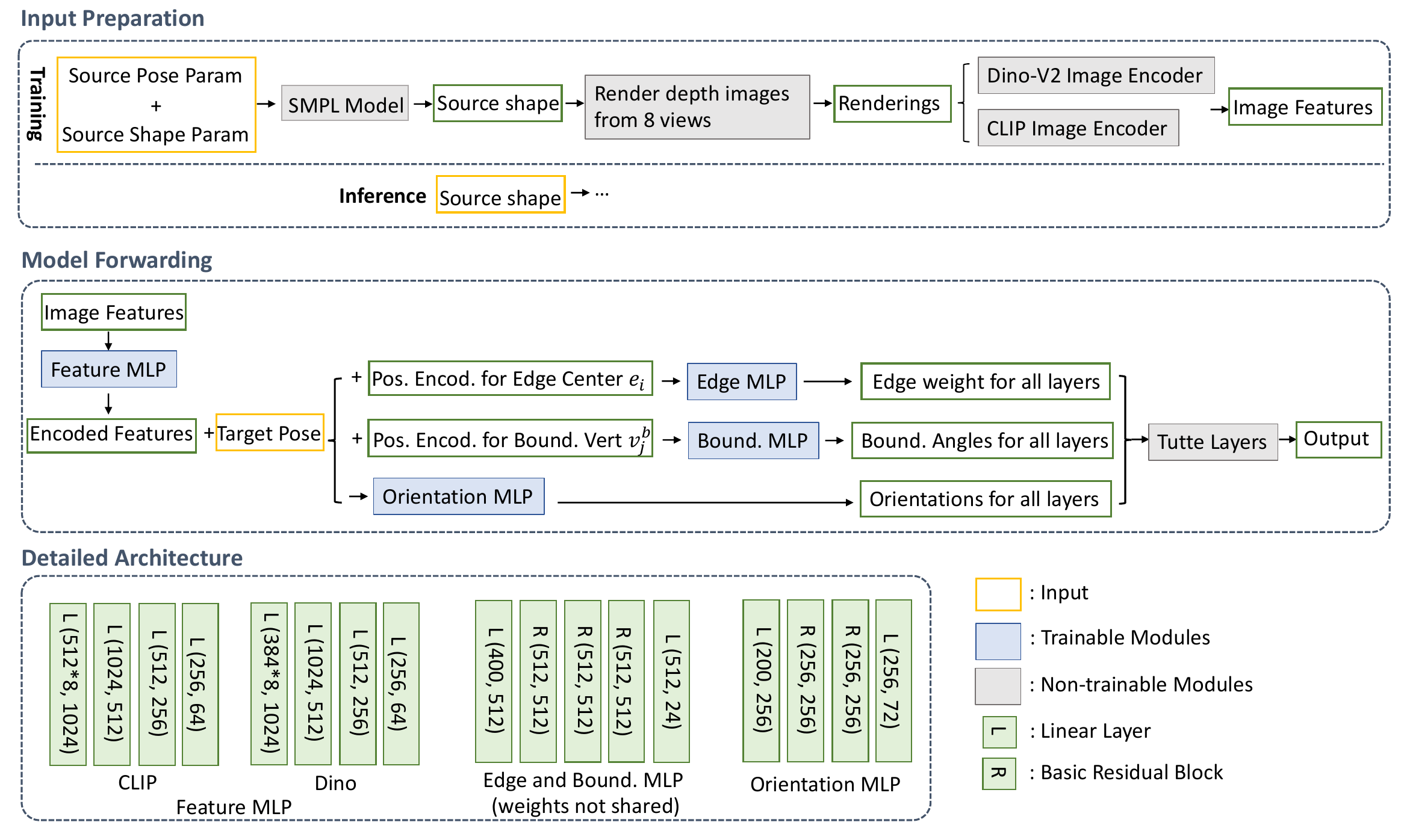}
\end{overpic}
\vspace{-0.1in}
\caption{Detailed model architecture and data preparation for the learning experiment (Section 4.2 in the main paper). }
\label{supp:Figure:architecture}
\end{figure*}

\noindent \textbf{Network Architecture of the Learning Experiment}.
The detailed process of data preparation and the model architecture are illustrated in Figure~\ref{supp:Figure:architecture}. 
For data preparation, we generate eight depth images for the input shapes and feed them into the CLIP~\cite{clip} and DINO-V2~\cite{dinov2} image backbones to extract image features. These features, along with the target pose parameters, serve as input for the deformation model.
In the comparisons presented in the main paper, we consistently set the source pose parameters to the canonical pose. The source shape parameters and target pose parameters are sampled following Section~\ref{supp:sec:data_learning}. During inference, the input is not necessarily restricted to the SMPL model. Instead, we directly take the 3D model as input and render images onto it.
During model forwarding, the image features are initially encoded in smaller feature vectors. These encoded features, along with the target pose parameters, act as conditioning vectors, guiding the prediction of Tutte parameters. Three networks are used to predict edge weights, boundary angles, and plane orientations. 
The edge MLP takes the positional encoding of each edge center, along with the conditioning vectors, as input and outputs the edge weights for all layers of each edge. Similarly, the boundary MLP takes the positional encoding of each boundary vertex position and the conditioning vectors as input, producing the boundary angle for all layers of each boundary vertex. The orientation MLP takes only the conditioning vector as input and outputs the orientations for all layers.
In our experiments, we set the number of layers to 24, the resolution of the mesh to $11\times 11$, and the positional encoding frequency to 50. Additional details on channel dimensions can be found in Figure~\ref{supp:Figure:architecture}

\section{Detailed Baseline Settings}

\subsection{NeRF Deformation Baselines (Section 4.1 in the main paper)}
\begin{itemize}
    \item \textbf{NeRF-Editing~\cite{nerfediting}} 
   We adhere to the procedures outlined in the official GitHub repository at \url{https://github.com/IGLICT/NeRF-Editing}. For the Lego data set, we utilize the checkpoint and cage data provided by the manufacturer. In the case of the Trex and Robot datasets, we follow the instructions on GitHub and receive direct guidance from the authors to train the model and generate the cage.
During the editing phase, we input their extracted mesh into our pre-optimized model, incorporating specified handle constraints to obtain the deformed mesh. Subsequently, we follow their prescribed steps to achieve the final rendering results.
    
    \item  \textbf{Deforming-nerf~\cite{deformingRFCages}}
    We closely follow the procedures outlined in the official GitHub repository at \url{https://github.com/xth430/deforming-nerf}. This method necessitates an initial deformation of the cage vertices, with subsequent harmonic coordinate interpolation employed to determine the corresponding deformed positions for ray points. Typically, users manually perform the deformation of the cage vertex. However, in our case, we lack explicit instructions on how to manipulate cage vertices to meet handle constraints.
Instead, we employ a different approach. Initially, our pre-optimized model is used to obtain the deformed positions for the shape mesh. Leveraging the differentiability of barycentric interpolation, we optimize the cage vertices so that their interpolation leads to the deformed shape positions. In this optimization process, we consider the cage vertices as variables subject to optimization. The procedure takes the undeformed shape points as input, utilizes the cage vertices to derive the deformed shape points, and optimizes the $L_2$ loss between the resulting points and the ground truth (GT) deformed points, those generated by our deformed model. We sample 10,000 points from the shape and iteratively optimize the $L_2$ loss until stability is reached and the loss is lower than $1 \times 10^{-5}$.
For the Lego and Robot datasets, the author has generously provided pre-trained models. However, we trained the Trex model from scratch following the provided instructions.
    
    \item  \textbf{SPIDR~\cite{liang2022spidr}}
   We adhere to the guidelines presented in the official GitHub repository available at \url{https://github.com/nexuslrf/SPIDR}. 
   For elastic deformation, we employ the notebook accessible at \href{https://github.com/nexuslrf/SPIDR/blob/main/deform_tools/examples/mesh_guided_deformation_ARAP.ipynb}{this link}.
   With handle constraints specified, the original method utilizes open3d for mesh deformation, a process that occasionally yields unsatisfactory outcomes due to non-injectivity issues. To ensure a fair comparison, we substitute the open3d deformation function with our pre-optimized model, seamlessly integrating it into the remaining steps outlined in their methodology. In particular, the checkpoints provided are exclusively available for the Lego and Trex datasets.
\end{itemize}

\subsection{Injective Baselines (Section 4.2 in the main paper)}
\begin{itemize}
    \item \textbf{i-ResNet~\cite{iresnet}} 
    We adopt the implementation provided at \url{https://github.com/stevenygd/NFGP}. 
    The chosen hyperparameters align with the configuration specified in the deformation settings, available at \href{https://github.com/stevenygd/NFGP/blob/master/configs/deformation/armidillo_LR_s1e-1_b1e-3.yaml}{this link}. Specifically, we configure the model with six layers, a positional encoding frequency of 5, and a latent dimension of 256. During the learning experiment, we condition the generation by appending the conditional feature vector to the positional encoding.
    
    \item  \textbf{RealNVP~\cite{realnvp}}
    We adopt the implementation available at \url{https://github.com/ikostrikov/pytorch-flows} and perform a thorough hyperparameter search to optimize performance. Regarding the mask selection, given our three-dimensional input and output, we employ an alternating approach across the three dimensions. This involves masking out one dimension at a time during each iteration to facilitate the RealNVP layer forward pass. Based on the best performance observed and considering the compatibility with our model size, we set the number of layers to 6 and the hidden dimension to 32.
    In the learning phase, we seamlessly integrate the conditional input, following the approach outlined in their code.
    
    \item  \textbf{NeuralODE~\cite{chen2018neural}}
   We align with the implementations available at \url{https://github.com/hjwdzh/MeshODE} and \url{https://github.com/maxjiang93/ShapeFlow}, both of which utilize NeuralODE for mesh deformation. To accommodate the size of our model, we employ four Linear layers with a latent dimension of 120.
During the learning experiment, we adhere to the ShapeFlow~\cite{jiang2020shapeflow} configuration, incorporating a conditional vector for every sample in the ODE function. We set the absolute and relative tolerance in odeint at $10^{-4}$.
In the elastic deformation experiment, we leverage the pytorch gradient function to invoke their built-in Jacobian computation in the ODE solver. Subsequently, using the same handle constraints, we perform deformation. To achieve optimal results and match our model size (24 layers with mesh resolution $25\times 25$), we set the latent dimension to 200 and employ the Adam optimizer for 12,000 steps with a learning rate of $10^{-3}$.
\end{itemize}

{
    \small
    \bibliographystyle{ieeenat_fullname}
    \bibliography{main}
}